\documentclass[10pt,twocolumn,letterpaper]{article}

\usepackage[pagenumbers]{cvpr} 

\usepackage{cvpr}
\usepackage{graphicx}
\usepackage{amsmath}
\usepackage{amssymb}
\usepackage{booktabs}
\usepackage{changepage}
\usepackage{multirow}
\usepackage{colortbl}
\usepackage{xcolor}
\usepackage{array}
\usepackage{minipage-marginpar}
\usepackage{bm}
\usepackage{lmodern} 
\usepackage[dvipsnames]{xcolor}
\usepackage{newtxtext} 

\newcommand{\parahead}[1]{\vspace{1.5mm}\noindent\textbf{#1}\ }
\newcommand{\ourmodel}{\textsc{Dynapo}\xspace}

\usepackage{amsmath,amssymb,amsfonts}
\usepackage{graphicx}
\usepackage{booktabs}
\usepackage{graphicx}
\usepackage{caption}

\newcommand{\video}{\mathbf{V}}
\newcommand{\realnum}{\mathbb{R}}

\newcommand{\maskseq}{\mathbf{M}}
\newcommand{\keycand}{\mathcal{C}}

\newcommand{\pB}{\mathbf{p}}

\newcommand{\rB}{\mathbf{r}}

\newcommand{\scless}{\resizebox{!}{.7ex}{\textless}}
\newcommand{\scgreater}{\resizebox{!}{.7ex}{\textgreater}}

\definecolor{cvprblue}{rgb}{0.21,0.49,0.74}
\usepackage[pagebackref,breaklinks,colorlinks,allcolors=cvprblue]{hyperref}

\def\paperID{9622} 
\def\confName{CVPR}
\def\confYear{2026}

\title{The Dynamic Prior: Understanding 3D Structures for Casual Dynamic Videos}

\author{
Zhuoyuan Wu$^1$\qquad
Xurui Yang$^2$\qquad
Jiahui Huang$^3$\qquad
Yue Wang$^4$\qquad
Jun Gao$^{3, 5}$ \\
$^1$PKU \ \ \
$^2$Independent Researcher  \ \ \ 
$^3$NVIDIA \ \ \ 
$^4$USC \ \ \ 
$^5$University of Michigan \ \ \ 
}

\begin{document}
\maketitle

\begin{abstract}
\noindent Estimating accurate camera poses, 3D scene geometry, and object motion from in-the-wild videos is a long-standing challenge for classical structure from motion pipelines due to the presence of dynamic objects. 
Recent learning-based methods attempt to overcome this challenge by training motion estimators to filter dynamic objects and focus on the static background. 
However, their performance is largely limited by the availability of large-scale motion segmentation datasets, resulting in inaccurate segmentation and, therefore, inferior structural 3D understanding. 
In this work, we introduce the Dynamic Prior (\ourmodel) to robustly identify dynamic objects without task-specific training, leveraging the powerful reasoning capabilities of Vision-Language Models (VLMs) and the fine-grained spatial segmentation capacity of SAM2.
\ourmodel can be seamlessly integrated into state-of-the-art pipelines for camera pose optimization, depth reconstruction, and 4D trajectory estimation.
Extensive experiments on both synthetic and real-world videos demonstrate that \ourmodel not only achieves state-of-the-art performance on motion segmentation, but also significantly improves accuracy and robustness for structural 3D understanding. \href{https://github.com/wuzy2115/DYNAPO}{Code} is available.

\end{abstract}
\vspace{-0.3cm}    
\vspace{-3mm}
\section{Introduction}
\label{sec:intro}
\vspace{-1mm}

Robustly estimating accurate 3D structures from real-world videos, including camera poses, scene geometry, and object motion, is a fundamental problem in computer vision, with applications across VR/AR, robotics, autonomous vehicles, and more broadly, spatial intelligence. 
With decades of research, classical methods, such as Structure from Motion (SfM)~\cite{4408933} and Simultaneous Localization And Mapping (SLAM)~\cite{slam}, have yielded mature algorithms for stationary scenes.
However, their reliance on the epipolar constraints poses significant challenges to dynamic scenarios with moving objects, which are pervasive in real-world videos.  


The fundamental problem is the ambiguity introduced by the moving objects, which violates the epipolar constraints.
Many recent works~\cite{zhang2024megasam,wimbauer2025anycam,jin2025stereo4d,chen2025back,teed2021droid,zhang2024monst3r,wang2025vggt,xiao2025spatialtracker,mcgann2022robust,huang2019clusterslam,judd2024multimotion} tackle this problem by training a neural network to identify the dynamic regions and filter them out in the bundle adjustment. However, these approaches are fundamentally limited by the availability of large-scale datasets for motion segmentation, and thus, produce inaccurate motion masks for in-the-wild videos (as shown in Fig.~\ref{fig:teaser}), leading to inferior 3D scene understanding results.
BA-Track~\cite{chen2025back} jointly optimizes static and dynamic scene geometry through a motion decoupled 3D tracking estimator model. Yet, training this estimator model is also limited by the availability of large-scale training data.

Large vision foundation models, such as VLMs~\cite{hurst2024gpt,comanici2025gemini,bai2025qwen2,wang2025internvl3}, SAM~\cite{ravi2024sam2}, and Depth Anything~\cite{yang2024depth}, has demonstrated unprecedented success in vision perception. While many recent 
methods~\cite{zhang2024megasam,wimbauer2025anycam,jin2025stereo4d,chen2025back,huang2025vipe} have started integrating the prior knowledge from these models to improve the robustness and accuracy; they are mainly limited to monocular depth estimation and tracking~\cite{yang2024depth,piccinelli2024unidepth,spann3r}.
In this work, we present the Dynamic Prior (\ourmodel), which leverages the powerful modern vision foundation model to overcome the generalization gap for motion segmentation, and in turn enhances structural 3D understanding.
Specifically, we study the combination of Vision-Language Models (VLMs)~\cite{hurst2024gpt, comanici2025gemini, bai2025qwen2, yang2025qwen3} and 2D segmentation models (SAM2)~\cite{ravi2024sam2}, where VLMs perform high-level reasoning to identify dynamic objects in the video and prompt SAM2 to segment out each dynamic object. Since both VLMs and SAM2 are pre-trained on vast internet-scale data, \ourmodel is robust and generalizable, and achieves state-of-the-art performance on motion segmentation.

We further integrate our 
\ourmodel in various state-of-the-art 3D structural understanding frameworks. \textbf{(I)} We integrate it into camera pose optimization pipelines~\cite{wimbauer2025anycam} to refine the camera poses by using our dynamic mask for filtering in bundle adjustment. Our framework significantly improves the accuracy of pose estimation for a wide range of methods~\cite{zhang2024megasam,wang2025vggt,wimbauer2025anycam,zhang2024monst3r,chen2025easi3r,xiao2025spatialtracker,wang2025continuous,chen2025ttt3r} on both synthetic and real-world videos. 
\textbf{(II)} We integrate it with the consistent depth optimization pipeline from MegaSaM~\cite{zhang2024megasam} for better recovery of 3D scene geometry. Using the dynamic mask from \ourmodel achieves significant improvement on the depth estimation across various datasets.
\textbf{(III)} We extend it with Stereo4D~\cite{jin2025stereo4d} for 4D trajectory estimation. By simply replacing the motion mask from Stereo4D with \ourmodel, the same optimization pipeline can be significantly improved for better trajectory estimation. 
These extensive applications and experiments demonstrate the effectiveness of motion estimation from our \ourmodel, moving a step towards robust and accurate structural 3D understanding.

\begin{figure*}
    \centering
    \includegraphics[width=1.0\textwidth]{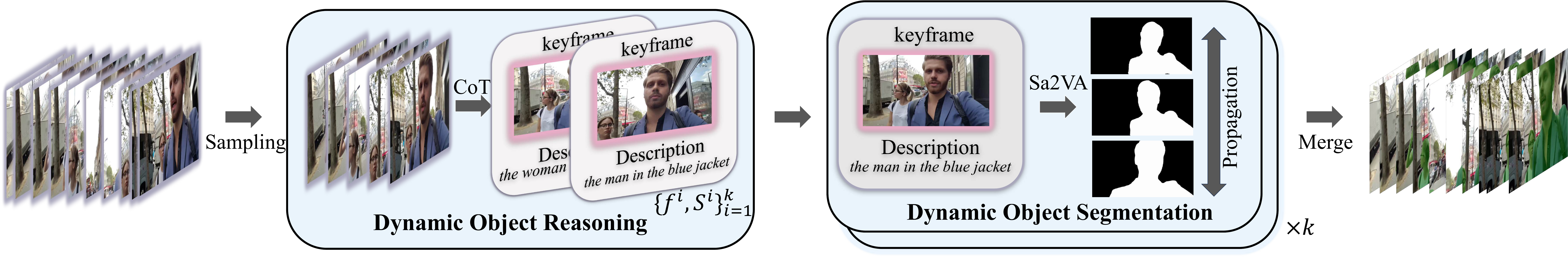}
    \vspace{-5mm}
    \caption{\footnotesize Overview of \ourmodel. Given a video sequence, the \textbf{Dynamic Object Reasoning} takes input of sub-sampled keyframes, reasons all the dynamic objects within the video, and generates descriptions $s^i$ and the frame number $f^i$ for each dynamic object (in total, $k$ objects). The \textbf{Dynamic Object Segmentation} then generates a mask sequence for each dynamic object, which we take an average to produce the final dynamic mask.
    }
    \label{fig:pipeline}
\end{figure*}

\section{Related Work}
\vspace{-0.1cm}
\subsection{Motion Segmentation}
\vspace{-1mm}
Motion segmentation aims to segment moving objects from static backgrounds. We broadly categorize prior work into flow-based and trajectory-based approaches. Flow-based methods~\cite{cho2023treating, xie2024moving, EM, OCLR, yang2019CIS} analyze short-term pixel displacements, and often fail with slow movements or significant camera motion. Trajectory-based methods~\cite{tokmakov2017learning, ochs2013segmentation, RCF, dutt2017fusionseg, lee2011key} track points over longer sequences, offering more robustness to occlusion but at a higher computational cost and sensitivity to tracking errors. 
More recently,
SegAnyMo~\cite{huang2025segment} combines long-range trajectories with semantic features and leverages SAM2~\cite{ravi2024sam2} for mask densification. Romo~\cite{golisabour2024} improves SfM by robustly combining optical flow and epipolar cues with a pre-trained
video segmentation model. However, both paradigms primarily rely on geometric cues and struggle to incorporate semantic context. This limitation prevents them from understanding complex dynamic scenes where an object's motion is defined by its interaction with the environment. On the contrary, our method leverages chain-of-thought reasoning for complex video tasks~\cite{kao2025thinkvideohighqualityreasoningvideo}, providing a deeper, context-aware understanding of dynamic scenes.

\vspace{-1mm}
\subsection{Dynamic Structure from Motion}
\vspace{-1mm}
Traditional SfM pipelines~\cite{Schoenberger2016CVPR, snavely2006photo, agarwal2011building, pollefeys2004visual} have excelled at reconstructing static scenes by matching local features across multiple views.
However, the presence of dynamic objects violates the constraints, often leading to catastrophic failures.
Recent advances in deep learning have led to two main paradigms for addressing this challenge. The first is feed-forward models~\cite{wang2024dust3r,wang2025vggt}, which directly regress camera parameters and 3D structures from images. While being powerful, these models, along with their follow-up works~\cite{spann3r,DUSt3Rpp,SLAM3R,MASt3R-SLAM}, are typically trained on static datasets and struggle with dynamic content~\cite{zhao2023pseudo}. 
To mitigate this, many methods~\cite{zhang2024monst3r,mast3r,wang2025continuous,daser,chen2025easi3r}
have explored explicitly handling moving objects by fine-tuning on dynamic scenes. However, they are limited by the availability of large-scale dynamic datasets.
The second paradigm integrates deep neural networks into SLAM systems. Models such as DROID-SLAM~\cite{teed2021droid}, MegaSAM~\cite{zhang2024megasam}, and LEAPVO~\cite{chen2024leap} build upon neural motion estimation to create robust systems that can operate in dynamic environments. Other methods like ParticleSfM~\cite{zhao2022particlesfm}, Robust-CVD~\cite{kopf2021rcvd} and CasualSAM~\cite{zhang2022structure} also aim to improve robustness by explicitly identifying and down-weighting the influence of dynamic elements during optimization. However, a common limitation is the accuracy of motion segmentation, which is limited by the training data and generalizes poorly for in-the-wild videos. Our work integrates the reasoning-based motion segmentation model into SfM, and significantly improves the performance of structural 3D understanding. 
\section{The Dynamic Prior}
\label{sec:dynamic_prior}
Our goal is to estimate camera poses, scene geometry and 4D trajectory from casual dynamic videos.
We first discuss the details of \ourmodel for motion segmentation using dynamic prior in this section, and integrate it into structural 3D understanding pipelines in Sec.~\ref{sec:3D_understand}.

As shown in Fig.~\ref{fig:pipeline}, with a video sequence $\video = \{I_t \in \realnum^{H \times W \times 3}\}_{t=1}^T$, \ourmodel estimates a  binary mask sequence $\{\maskseq_t \in \{0,1\}^{ H \times W}\}_{t=1}^T$, capturing all the dynamic objects in the video. \ourmodel achieves this through a two-stage framework, each leveraging prior knowledge from pretrained vision foundation models. In the first stage (Sec.~\ref{sec:dynamic_obj_reason}), we prompt a VLM to reason about the dynamic objects in the video, identify the keyframe where the object is most salient, and generate descriptions for each dynamic object. The descriptions and keyframes are then fed into the second stage (Sec.~\ref{sec:dynamic_obj_seg}), which uses Multi-Modal Large Language Model (MLLM) with SAM2 to generate the dynamic mask. The whole process is fully automatic without post-training and is applicable to in-the-wild videos. 

\subsection{Dynamic Object Reasoning}
\label{sec:dynamic_obj_reason}
The key question for reasoning dynamic objects is to understand \textit{``which objects are moving"} in the video. The VLMs, such as GPT-4o~\cite{hurst2024gpt}, Gemini2.5~\cite{comanici2025gemini}, and Qwen-VL~\cite{bai2025qwen2,yang2025qwen3}, have been trained on various video understanding tasks with emergent reasoning capabilities~\cite{hurst2024gpt}. We design a prompting mechanism to let VLMs explicitly reason about dynamic objects that are most representative in the video.

First, to reduce token counts and improve efficiency, we uniformly subsample a set of keyframes from the video, $\keycand = \{I_t\}, {t \in \mathcal{T}_{cand}}$, where $\mathcal{T}_{cand}$ is the set of frame indices for keyframes. 
The VLMs are then prompted through a structured chain-of-thought process to analyze the keyframes $\keycand$ and reason about object motion. Specifically, \textbf{(i).} VLMs analyze each keyframe and identify moving objects across the frames. \textbf{(ii).} For each object, VLMs determine the occlusion, and start and end frames if object appears and disappears. \textbf{(iii).} VLMs finally generate the frame number where the object is most salient and a description for each object. 
The output is $k$ tuples corresponding to $k$ unique dynamic objects $\{(f^i, s^i)\}_{i=1}^k$, where $f^i \in \keycand$ is the selected frame for the $i$-th dynamic object, and $s^i$ is a description for the object's location and appearance within the frame $f^i$. Further details are provided in the Appendix. 


\subsection{Dynamic Object Segmentation}
\label{sec:dynamic_obj_seg}
With the identified dynamic objects and their descriptions, we first generate the mask for each object in the corresponding keyframe, and then propagate the mask from the keyframe into the whole video. The final binary mask $\{\maskseq_t \in \{0,1\}^{ H \times W}\}_{t=1}^T$ is obtained by averaging across all the dynamic object. We detail each step below:


\parahead{Mask Segmentation at Keyframe.}
For each identified dynamic object $i$, we generate a binary mask $\mathbf{m}^i \in \{0,1\}^{H \times W}$ on the corresponding keyframe $f^i$, leveraging the description $s^i$.
Our segmentation pipeline leverages pretrained Sa2VA~\cite{sa2va}, which synergizes a multi-modal large language model (MLLM) with SAM2~\cite{ravi2024sam2} decoder. We first encode both text and visual inputs into a unified token embedding space and process it through an MLLM, which generates specialized instruction tokens (i.e., ``[SEG]'' tokens) alongside conventional language outputs. The instruction token serves as a prompt for the SAM2 decoder, directing it to produce the segmentation masks for the selected dynamic object. 

\parahead{Mask Propagation from Keyframe to Video.}
The mask $\mathbf{m}^i$ at the keyframe needs to propagate to the entire video sequence to form the complete mask sequence $ \{\mathbf{m}^i_t \in \{0,1\}^{ H \times W}\}_{t=1}^T$ for the dynamic object $i$. We leverage the memory mechanism in SAM2~\cite{ravi2024sam2}, without modification. We provide a rough description in the Appendix and refer the readers to the original paper for details.
With this memory mechanism, SAM2 can generate a temporally coherent mask throughout the entire video sequence. 
Applying the same process to for all $k$ identified dynamic instances, we obtain $k$ separate mask sequences, $ \{\mathbf{m}^1_t \}_{t=1}^T, \cdots,  \{\mathbf{m}^{k}_t\}_{t=1}^T$. 

\parahead{Mask Merging.}
We merge all the instance-level video mask sequences into a unified video mask for the final output by simply computing the element-wise union of all individual instance masks:
    $\maskseq_t = \bigcup_{i=1}^{k} \mathbf{m}^i_t.$
This process generates the final binary mask sequence $\{\maskseq_t\}_{t=1}^T$, where each mask identifies all dynamic regions in its corresponding frame.

\begin{figure}[!t]
    \vspace{-1mm}
    \centering
    \includegraphics[width=1\linewidth]{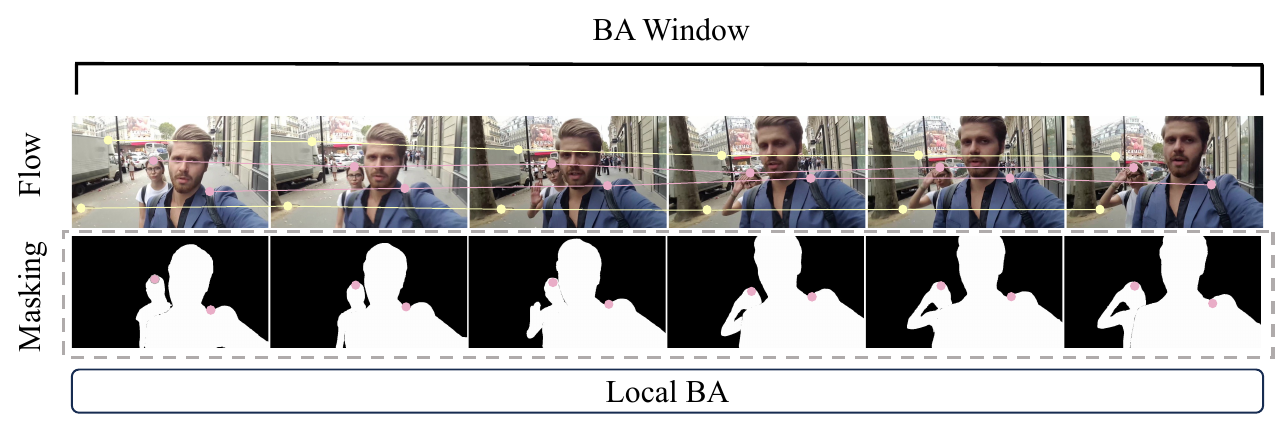}
    \vspace{-3mm}
    \caption{Illustration of our BA. When calculating the reprojection loss, 
    we effectively remove the dynamic objects (pink tracks) from the loss function using the mask from \ourmodel, and the BA can only focus on the static background for optimization.
    }
    \label{fig:BA}
    \vspace{-0.2cm}
\end{figure}

\section{3D Structure Understanding from Videos}
\label{sec:3D_understand}
The estimated dynamic masks from \ourmodel (Sec.~\ref{sec:dynamic_prior}) serve as the filter for the dynamic object, from where we can optimize the camera pose, scene geometry, and 4D tracking using conventional methods. We integrate \ourmodel into three state-of-the-art structural 3D understanding frameworks.  We first integrate into camera pose optimization pipelines~\cite{wimbauer2025anycam} to refine the camera poses with bundle adjustment in Sec.~\ref{subsec:bundle_adjustment}. We then integrate it with MegaSam~\cite{zhang2024megasam} for consistent depth optimization in Sec.~\ref{subsec:depth_optimization} to recover 3D scene geometry. Finally, we extend it with Stereo4D~\cite{jin2025stereo4d} for accurate 4D trajectories estimation in Sec.~\ref{subsec:track_optimization}.

\subsection{Camera Pose Optimization}
\label{subsec:bundle_adjustment}

\parahead{Problem Setting.}
We incorporate \ourmodel into the bundle adjustment (BA) process from AnyCam~\cite{wimbauer2025anycam} to optimize camera poses. This process acts as a plug-and-play post-processor that can be applied to refine the camera pose estimation from various methods.
The inputs are the initial camera poses $\{\mathbf{P}_t\}$, intrinsics matrix $K$, and depth maps $\{\mathbf{D}_t\}$, and dense optical flow fields $\mathbf{F}^{t \to t+1}$ between adjacent frames from any off-the-shelf 3D estimation model. We optimize the camera pose with the dynamic object mask derived from \ourmodel. 



Let $\pi_K(\mathbf{x}): \mathbb{R}^3 \rightarrow \mathbb{R}^2$ denote the projection function that maps a 3D point $\mathbf{x}$ in the camera's coordinate system to a 2D pixel $\mathbf{p}$ on the image plane, parameterized by the intrinsics $K$. 
Accordingly,  $\pi_K^{-1}(\mathbf{p}, d): (\mathbb{R}^2, \mathbb{R}_+) \rightarrow \mathbb{R}^3$ denotes unprojecting a pixel $\mathbf{p}$ with depth $d$ into 3D. Following AnyCam~\cite{wimbauer2025anycam}, we sample  a grid of points in a starting frame $t_0$ and track them for 8 frames by chaining the dense optical flows $\mathbf{F}^{t_0 \to t_0+1},\cdots, \mathbf{F}^{t_0+1 \to t_0+8}$,
forming a track: $((\mathbf{p}_1, \ldots, \mathbf{p}_8), (m_1, \ldots, m_8), d_1)$, where $\{\mathbf{p}_i\}$ are the pixel locations, $\{m_i\}$ are the corresponding binary motion mask values, and $d_1$ is the depth of the first point. We take the mask value from our generated motion mask from Sec.~\ref{sec:dynamic_prior}: 
\begin{equation}
    m_i = 1 - \maskseq_{t_0+i-1}(\mathbf{p}_i),
\end{equation}
where $\maskseq_{t_0+i-1}(\mathbf{p}_i)$ denotes the mask value at pixel location $\mathbf{p}_i$.
The initial depth is taken from the input depth map.

\parahead{Optimization.}
We jointly optimize camera poses, intrinsics and depths of anchor points to minimize the reprojection error on the static regions, indicated by our motion masks. Specifically, the loss for a single track is:
\begin{equation}
    \mathcal{L}_{\mathsf{Repr}} = \sum_{i=2}^{8}  m_i \cdot \left\lVert \pi_K(\mathbf{P}^{1\rightarrow i} \pi_K^{-1}(\mathbf{p}_1, d_1)) - \mathbf{p}_i\right\rVert_1,
\end{equation}
The term $m_i$ ensures that the reprojection error only focuses on static points when $m_i=0$, as shown in Fig.~\ref{fig:BA}. 

To ensure a plausible camera trajectory and prevent jerky movements, we additionally incorporate a smoothness regularizer. This term penalizes deviations from a constant-velocity camera motion by minimizing the difference between consecutive camera poses:
\begin{equation}
    \mathcal{L}_{\mathsf{smooth}} = \sum_{t=1}^{6} \left\lVert \left(\mathbf{P}^{t\rightarrow t+1}\right)^{-1} \mathbf{P}^{t+1\rightarrow t+2} - \mathbf{I}_4\right\rVert_{1,1},
\end{equation}
where $\mathbf{I}_4$ is the $4\times 4$ identity matrix. The final optimization objective is a weighted sum of the reprojection error and the smoothness term:
\begin{equation}
    \mathcal{L}_\mathsf{BA} = \mathcal{L}_{\mathsf{Repr}} + \lambda_\mathsf{smooth}\mathcal{L}_{\mathsf{smooth}}.
\end{equation}

\begin{figure}[!t]
    \centering
    \vspace{-1mm}
    \includegraphics[width=1\linewidth]{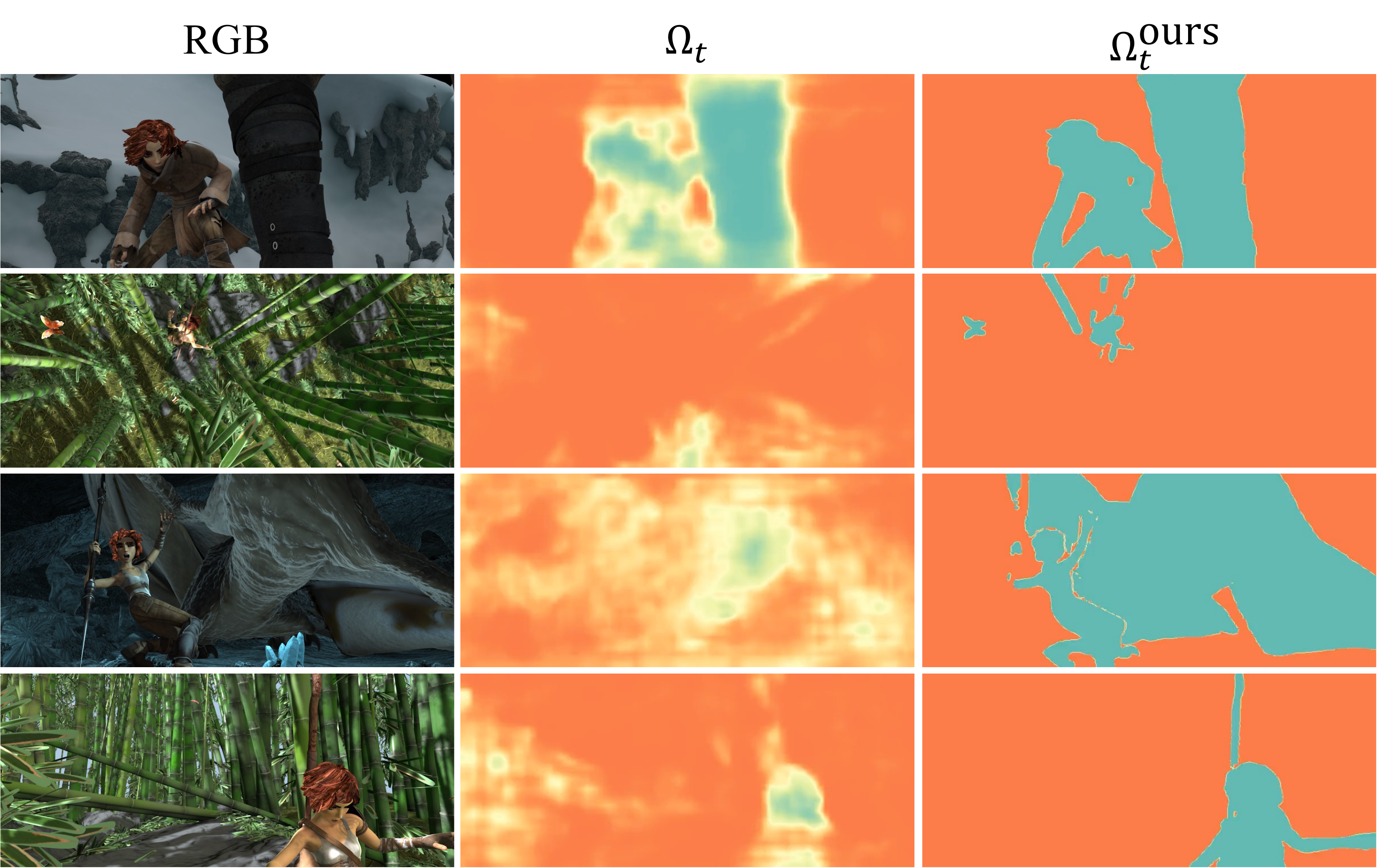}
    \vspace{-3mm}
    \caption{Visualization of uncertainty map. MegaSaM produces inaccurate dynamic masks, deteriorating depth optimization, while our \ourmodel generates cleaner and more plausible dynamic masks.}
    \label{fig:cvd_map}
    \vspace{-0.2cm}
\end{figure}

\subsection{Depth Optimization}
\label{subsec:depth_optimization}
\parahead{Background.}
We integrate \ourmodel with the Consistent Video Depth (CVD) optimization method from MegaSaM~\cite{zhang2024megasam} to optimize depth predictions. We briefly introduce it here and refer readers to the original paper for further details.
The objective in CVD is a weighed sum of pairwise 2D flow reprojection loss $\mathcal{L}_{\mathsf{flow}}$,  temporal depth consistency loss $\mathcal{L}_{\mathsf{temp}}$, and mono-depth prior loss $\mathcal{L}_{\mathsf{prior}}$: 
\begin{equation}
    \mathcal{L}_{\mathsf{cvd}} = \lambda_{\mathsf{flow}} \mathcal{L}_{\mathsf{flow}} +  \lambda_{\mathsf{temp}} \mathcal{L}_{\mathsf{temp}} + \lambda_{\mathsf{prior}} \mathcal{L}_{\mathsf{prior}}.
    \label{eq:depth_loss}
\end{equation}
To handle dynamic objects, CVD jointly optimizes an uncertainty map $\{\bm{\Omega}_t\}_{t=1}^T$. Specifically,
for a pixel $\mathbf{p}$ at frame $t$, its projected location at frame $j$ is computed as $\mathbf{p}' = \pi_K(\mathbf{P}^{t \to j} \pi_K^{-1}(\mathbf{p}, d)))$. The reprojection loss 
$\mathcal{L}_{\mathsf{flow}}$ computes the difference with the optical flow $\mathbf{F}^{t \to j}(\mathbf{p})$, weighted by the uncertainty $\bm{\Omega}_t(\mathbf{p})$:
{\footnotesize\begin{equation}
    \mathcal{L}_{\mathsf{flow}}^{i \to j} = \bm{\Omega}_i(\mathbf{p}) \cdot || \mathbf{p}' - (\mathbf{p} + \mathbf{F}^{i \to j}(\mathbf{p})) ||_1 + \log(1/\bm{\Omega}_i(\mathbf{p})),
    \label{eq:flow}
\end{equation}}
This loss encourages pixel disparity to be temporally consistent according to the estimated 2D optical flow. The uncertainty map $\{\bm{\Omega}_t\}_{t=1}^T$ is jointly optimized in the CVD pipeline. 

\parahead{Our Formulation.}
MegaSAM relies on the optimized uncertainty map $\{\bm{\Omega}_t\}_{t=1}^T$ to down-weight dynamic regions during optimization, which are typically inaccurate and prone to error.
We replace it with our semantics-aware motion mask, $\maskseq_t$ derived from Sec.~\ref{sec:dynamic_prior}, by simply initializing and finetuning it during optimization.
\begin{equation}
    \bm{\Omega}^{\text{ours}}_t = 1 - \maskseq_t.
\end{equation}
This substitution is applied to both $\mathcal{L}_{\mathsf{flow}}$ and $\mathcal{L}_{\mathsf{temp}}$, which allows them to completely disregard pixels on moving objects and focus exclusively on the static parts of the scene. A visual comparison is shown in Fig.~\ref{fig:cvd_map}. 

\begin{figure}[!t]
    \centering
    \includegraphics[width=1\linewidth]{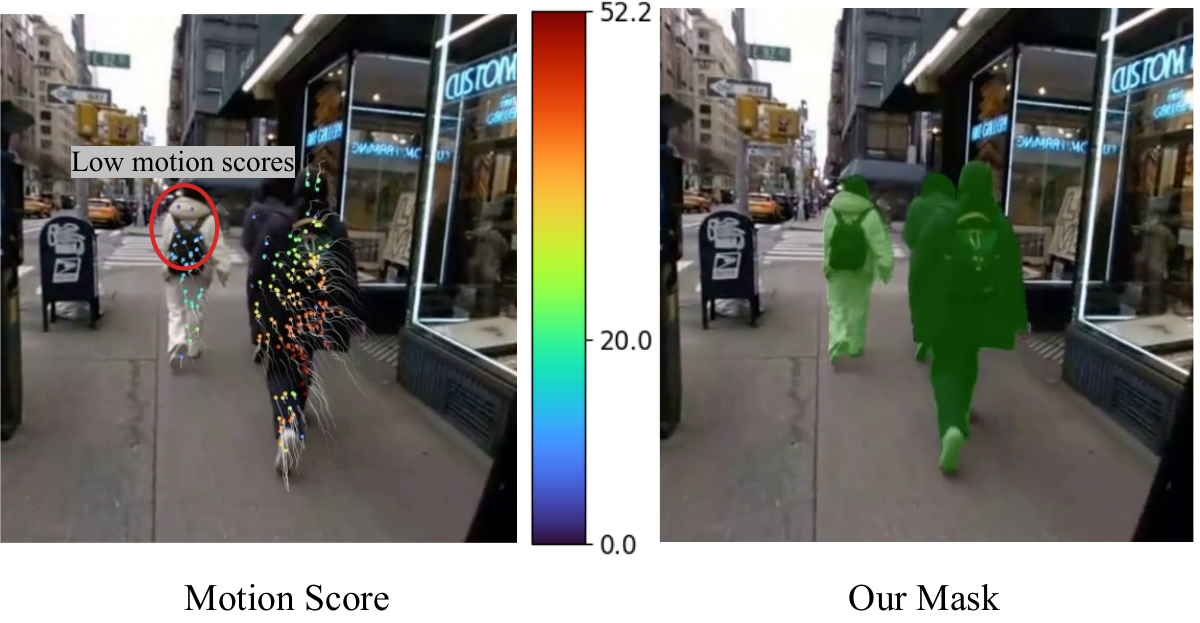}
    \vspace{-3mm}
    \caption{Visualization of motion mask.  Motion scores from Stereo4D~\cite{jin2025stereo4d} are inaccurate, where some dynamic tracks can have extremely small motion scores, leading to inappropriate loss weighting during optimization. 
    }
    \label{fig:motion_score}
    \vspace{-0.3cm}
\end{figure}


\subsection{4D Track Optimization}
\label{subsec:track_optimization}

\parahead{Background.}
Stereo4D~\cite{jin2025stereo4d} optimizes 4D trajectories from dynamic videos. 
For a 2D point track $\{\mathbf{p}_i\}_{i=1}^N$, with corresponding initial depth estimates $\{d_i\}_{i=1}^N$, the corresponding 3D points $\{\mathbf{x}_i\}_{i=1}^N$ can be obtained through unprojection $\mathbf{x}_i = \pi_K^{-1}(\mathbf{p}_i, d_i)$. The core idea of Stereo4D~\cite{jin2025stereo4d} is to optimize a per-frame scalar offset $\delta_i$ for each point along its corresponding camera ray, $\rB_i = (\mathbf{x}_i - \mathbf{c}_i) / ||\mathbf{x}_i - \mathbf{c}_i||$, where the camera center $\mathbf{c}_i$ is the translation component of the camera-to-world pose $\mathbf{P}_i$. The refined 3D position is then $\mathbf{x}'_i = \mathbf{x}_i + \delta_i \rB_i$. The main objective  in Stereo4D~\cite{jin2025stereo4d} is a combination of three terms:
\begin{equation}
\min_{\{\delta_i\}_{i=1}^N} \sigma(\mu)\mathcal{L}_{\mathsf{static}} + (1-\sigma(\mu))\mathcal{L}_{\mathsf{dynamic}} + \mathcal{L}_{\mathsf{reg}},
\label{eq:stereo4d}
\end{equation}
where $\mathcal{L}_{\mathsf{static}}$ encourages stationary points to remain fixed in world space, $\mathcal{L}_{\mathsf{dynamic}}$ promotes smooth trajectories for moving points, and $\mathcal{L}_{\mathsf{reg}}$ ensures consistency to the initial depth estimates.
$\sigma(\mu)$ is the function of  motion score $\mu$, defined as $\sigma(\mu) = \frac{1}{1 + \exp(\mu - \mu_0)}$, with $\mu_0 = 20$ as a threshold. 

While the formulation in Stereo4D~\cite{jin2025stereo4d} provides a solid foundation for track optimization, the objective heavily relies on the accuracy of the motion score. 
In the original paper, $\mu$ is defined as the as the $90^{th}$ percentile
of the track’s trail length across all frames:
\begin{equation}
\mu = \mathsf{Percentile}_{i=1:N}^{90}\left[\max_{w=1:w_o}\|\pi_i(\pB_i) - \pi_i(\pB_{i-w})\|\right]
\label{eqn:trail_length_def}
\end{equation}
Such a definition is prone to noise from camera ego-motion, misclassifying points for motion and degrading trajectory quality, as shown in Fig.~\ref{fig:motion_score}.

\parahead{Our Formulation.}
Our key modification is to replace this heuristic motion score with a more robust, semantics-aware signal derived directly from our \ourmodel.
Specifically, we determine a single motion score for each track based on dynamic masks for all the 2D points along the track. 
\begin{equation}
    \mu^{\text{ours}} = 
    \begin{cases}
      \mu_{dyn} & \text{if } \exists i \in \{1,\dots,N\} \text{ s.t. } \maskseq_{t_0+i-1}(\mathbf{p}_{i}) = 1 \\
      \mu_{stat} & \text{otherwise}
    \end{cases}
\end{equation}
where $\mathbf{p}_{i}$ is the $i$-th point of track and we simply set $\mu_{dyn}=25.0$ and $\mu_{stat}=15.0$. This approach creates a decisive, binary switch for each track, pushing the sigmoid weighting function $\sigma(\mu^{\text{ours}})$ to its extremes. This allows the objective Eq.~\ref{eq:stereo4d} to reasonably distinguish static or dynamic tracks, leading to significantly cleaner and more plausible 4D trajectories.

\begin{table}[t]
    \centering
    \scriptsize
\setlength{\tabcolsep}{5pt}
\renewcommand{\arraystretch}{1} 
\begin{tabular}{@{}l c c c@{}}
\toprule
 \multirow{2}{*}{Methods} & SegTrackv2 & FBMS-59 & DAVIS-2016 \\
 \cmidrule(l){2-2} \cmidrule(l){3-3} \cmidrule(l){4-4}
 & $\mathcal{J} \uparrow$ & $\mathcal{J} \uparrow$ & $\mathcal{J} \uparrow$ \\ \midrule

 CIS~\cite{yang2019CIS} & 62.0  & 63.6 & 70.3\\
 SIMO~\cite{Lamdouar2021SegmentingIM} & 62.2 & - & 67.8 \\
 OCLR-flow~\cite{OCLR} & 67.6  & 65.5 & 72.0\\
 OCLR-TTA~\cite{OCLR} & 72.3  & 69.9 & 80.8\\
 EM~\cite{EM} & 55.5  & 57.9 & 69.3\\
 RCF-Stage1~\cite{RCF} & 76.7  & 69.9 & 80.2\\
 RCF-All~\cite{RCF} & 79.6  & 72.4 & 82.1\\
 STM~\cite{10204532} & 55.0 & 59.2 & 73.2\\
 ABR~\cite{xie24appearrefine} & 76.6  & 81.9 & 71.8\\
 LRTR~\cite{karazija24learning} & 81.2 & 79.6 & 82.2\\
 SegAnyMo~\cite{huang2025segment} & 76.3  & 76.4 & 81.9\\
 RoMo~\cite{golisabour2024} & 67.7  & 75.5 & 77.3\\
 \midrule
 Ours & \textbf{83.3} & \textbf{84.0} & 81.6\\
\bottomrule
\end{tabular}
\label{tab:mos}
    \caption{Quantitative comparison on motion segmentation. Our method achieves state-of-the-art results in SegTrackv2 and FBMS-59, with comparable performance on the DAVIS-2016 dataset.}
    \label{tab:mos}
    \vspace{-3mm}
\end{table}

\section{Experiments}

We first evaluate the performance of \ourmodel for dynamic object segmentation in Sec.~\ref{exp:motion_seg}.
The improved dynamic object masks are, in turn, applied to pose optimization, depth optimization, and 4D track optimization for evaluation in Sec.~\ref{exp:pose_optimization}, Sec.~\ref{exp:depth_optimization}, and Sec.~\ref{exp:track_optimization}, respectively.

\subsection{The Dynamic Prior}
\label{exp:motion_seg}
\noindent{\bf Implementation Details.} We use  GPT-4o~\cite{hurst2024gpt} (version ``2024-05-13'') as our dynamic object reasoner.
The temperature of GPT-4o is set to $0.5$, and the maximum number of output tokens is $4096$.
For the segmentation, we use Sa2VA-8B~\cite{sa2va} and SAM2~\cite{ravi2024sam2} with the official pre-trained checkpoints.

\parahead{Baselines.} We compare our method against state-of-the-art approaches for moving object segmentation including SegAnyMo~\cite{huang2025segment}\footnote{we evaluate with the official code and checkpoints from \url{https://github.com/nnanhuang/SegAnyMo}}, RoMo~\cite{golisabour2024}, ABR~\cite{xie24appearrefine}, CIS~\cite{yang2019CIS}, EM~\cite{EM}, RCF~\cite{RCF}, OCLR~\cite{OCLR}, STM~\cite{10204532}, SIMO~\cite{Lamdouar2021SegmentingIM} and LRTR~\cite{karazija24learning}. 

\begin{figure}
    \centering
    \includegraphics[width=\linewidth]{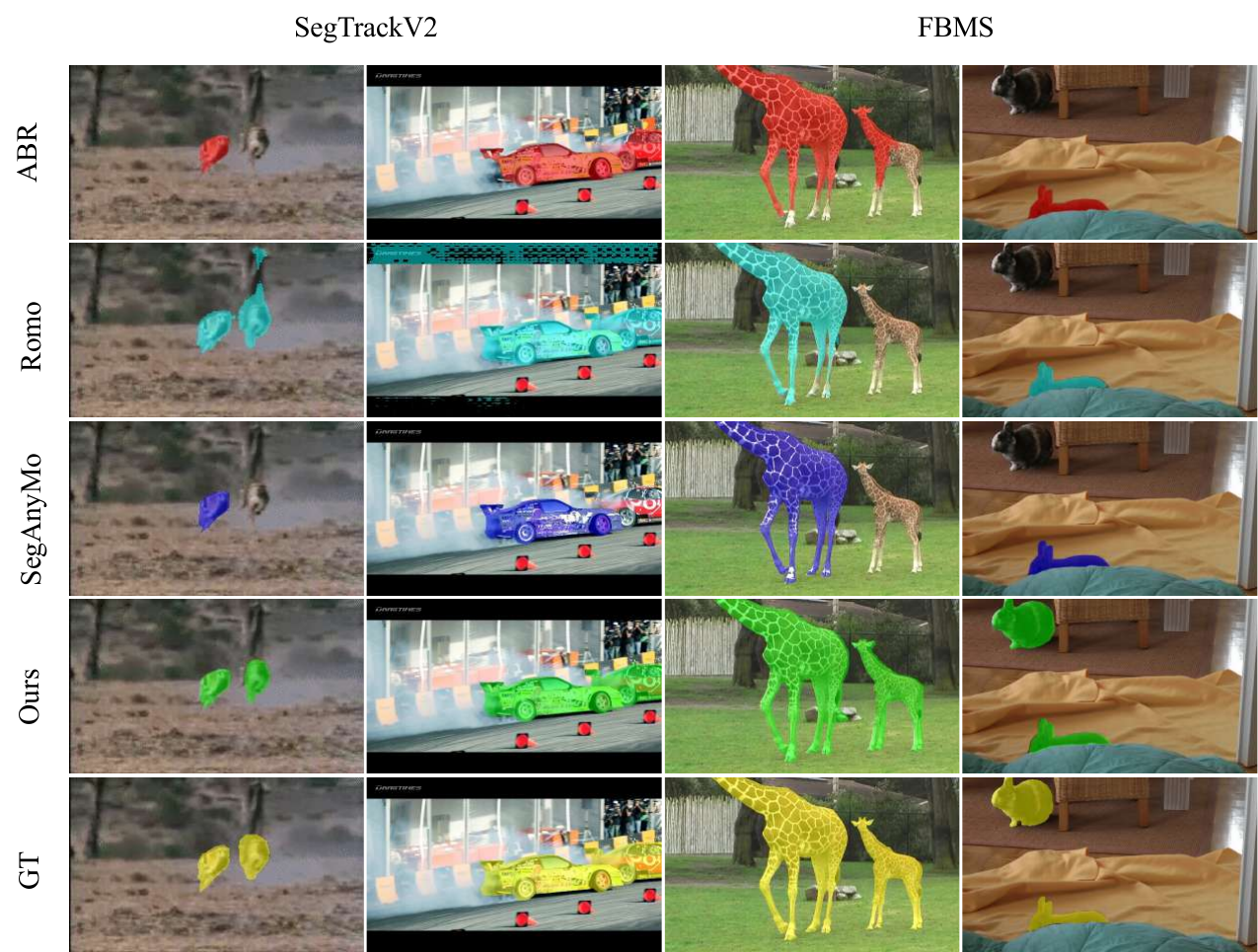}
    \vspace{-3mm}
    \caption{Qualitative results on motion segmentation. \ourmodel generates precise masks for all the dynamic objects, while baselines can neglect some dynamic objects or generate inaccurate masks.}
    \label{fig:qualitative}
\end{figure}
\parahead{Evaluation Datasets and Metrics.}
We evaluate on three well-established benchmarks for dynamic object segmentation: DAVIS-2016~\cite{Huang_2016}, SegTrackv2~\cite{li2013segtrackv2} and FBMS-59~\cite{ochs2014fbms59}. 
We follow prior work and compute
the region similarity ($\mathcal{J}$), also known as the Jaccard index, measures the overlap between the predicted and ground-truth masks.
For scenarios with multiple objects, we consolidate all foreground objects into a single mask for evaluation following prior work~\cite{dutt2017fusionseg,yang2019CIS,xie24appearrefine,OCLR}.

\parahead{Experimental Results.}
We provide quantitative comparisons in Table~\ref{tab:mos} with qualitative examples in Fig.~\ref{fig:qualitative}. Our approach achieves state-of-the-art results on FBMS-59 and SegTrackv2, with comparable performance on DAVIS-2016 
The significant improvement gains on both SegTrackv2 and the more challenging FBMS-59 benchmark highlight the robustness and accuracy of our proposed dynamic object reasoning in identifying moving objects.

\parahead{Ablation Studies.} Due to limited space, we provide results in the appendix. We ablate key components of our method, including the choice of Vision Language Models (VLMs) for dynamic scene reasoning, different backbones for our segmentation module, as well as the impact of the video frame sampling rate. 

\begin{figure}[!t]
    \centering
    \includegraphics[width=1\linewidth]{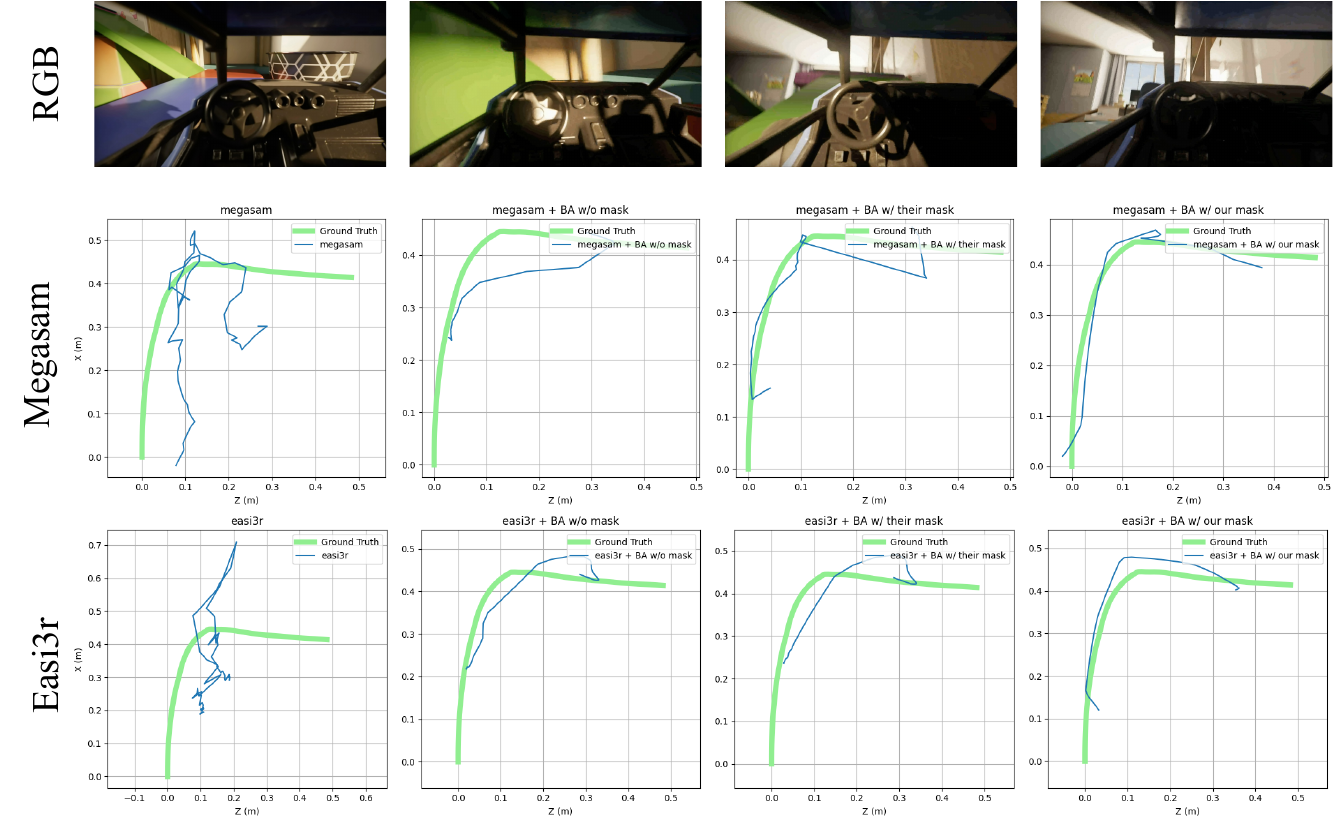}
    \caption{Visualization of estimated camera trajectories.}
    \label{fig:pose_opt}
    \vspace{-4mm}
\end{figure}

\subsection{Pose Optimization}
\label{exp:pose_optimization}
We evaluate the performance of our pose optimization by refining the camera poses outputs from a wide range of state-of-the-art SfM methods and optimizing their poses. 

\begin{table*}[!t]
\label{tab:ba_lightspeed}
\centering
\scriptsize
\caption{Quantitative comparison for camera pose estimation with different baselines on LightSpeed~\cite{rockwell2025dynamic} dataset.}
\setlength{\tabcolsep}{5pt}
\renewcommand{\arraystretch}{1.2}
\definecolor{GroupLeft}{RGB}{240,248,255}
\definecolor{GroupRight}{RGB}{255,240,245}

\begin{minipage}[t]{0.4\textwidth} 
\centering
\begin{tabular}{@{}l c c c@{}}
\toprule
\multicolumn{4}{@{}c}{\textbf{Methods that predict motion masks}} \\
\cmidrule(lr){1-4}
Method & ATE $\downarrow$ & RTE $\downarrow$ & RRE $\downarrow$ \\
\midrule

\rowcolor{GroupLeft}
MegaSam~\cite{zhang2024megasam} & 0.073 & 0.032 & 0.773 \\
\rowcolor{GroupLeft}
+ BA w/o mask & 0.090 \textcolor{red}{(+0.017)} & 0.028 \textcolor{PineGreen}{(-0.004)} & 1.003 \textcolor{red}{(+0.230)} \\
\rowcolor{GroupLeft}
+ BA w/ their mask & 0.072 \textcolor{PineGreen}{(-0.001)} & 0.019 \textcolor{PineGreen}{(-0.013)} & 2.504 \textcolor{red}{(+1.731)} \\
\rowcolor{GroupLeft}
+ BA w/ our mask & 0.052 \textcolor{PineGreen}{(-0.021)} & 0.019 \textcolor{PineGreen}{(-0.013)} & 0.910 \textcolor{red}{(+0.137)} \\

\rowcolor{GroupRight}
AnyCam~\cite{wimbauer2025anycam} & 0.139 & 0.032 & 1.101 \\
\rowcolor{GroupRight}
+ BA w/o mask & 0.133 \textcolor{PineGreen}{(-0.006)} & 0.028 \textcolor{PineGreen}{(-0.004)} & 1.057 \textcolor{PineGreen}{(-0.044)} \\
\rowcolor{GroupRight}
+ BA w/ their mask & 0.131 \textcolor{PineGreen}{(-0.008)} & 0.025 \textcolor{PineGreen}{(-0.007)} & 0.914 \textcolor{PineGreen}{(-0.187)} \\
\rowcolor{GroupRight}
+ BA w/ our mask & 0.125 \textcolor{PineGreen}{(-0.014)} & 0.025 \textcolor{PineGreen}{(-0.007)} & 0.884 \textcolor{PineGreen}{(-0.217)} \\

\rowcolor{GroupLeft}
Monst3r~\cite{zhang2024monst3r} & 0.112 & 0.029 & 0.998 \\
\rowcolor{GroupLeft}
+ BA w/o mask & 0.096 \textcolor{PineGreen}{(-0.016)} & 0.023 \textcolor{PineGreen}{(-0.006)} & 0.962 \textcolor{PineGreen}{(-0.036)} \\
\rowcolor{GroupLeft}
+ BA w/ their mask & 0.093 \textcolor{PineGreen}{(-0.019)} & 0.024 \textcolor{PineGreen}{(-0.005)} & 0.969 \textcolor{PineGreen}{(-0.029)} \\
\rowcolor{GroupLeft}
+ BA w/ our mask & 0.092 \textcolor{PineGreen}{(-0.020)} & 0.023 \textcolor{PineGreen}{(-0.006)} & 0.935 \textcolor{PineGreen}{(-0.063)} \\

\rowcolor{GroupRight}
Easi3r~\cite{chen2025easi3r} & 0.214 & 0.058 & 2.153 \\
\rowcolor{GroupRight}
+ BA w/o mask & 0.135 \textcolor{PineGreen}{(-0.079)} & 0.039 \textcolor{PineGreen}{(-0.019)} & 1.076 \textcolor{PineGreen}{(-1.077)} \\
\rowcolor{GroupRight}
+ BA w/ their mask & 0.150 \textcolor{PineGreen}{(-0.064)} & 0.038 \textcolor{PineGreen}{(-0.020)} & 1.079 \textcolor{PineGreen}{(-1.074)} \\
\rowcolor{GroupRight}
+ BA w/ our mask & 0.118 \textcolor{PineGreen}{(-0.096)} & 0.031 \textcolor{PineGreen}{(-0.027)} & 1.027 \textcolor{PineGreen}{(-1.126)} \\

\bottomrule
\end{tabular}
\label{tab:ba-left}
\end{minipage}
\hspace{0.13\textwidth} 
\begin{minipage}[t]{0.4\textwidth} 
\centering
\begin{tabular}{@{}l c c c@{}}

\toprule
\multicolumn{4}{@{}c}{\textbf{Methods that do not predict motion masks}} \\
\cmidrule(lr){1-4}
Method & ATE $\downarrow$ & RTE $\downarrow$ & RRE $\downarrow$ \\
\midrule

\rowcolor{GroupLeft}
SpatialTrackerv2~\cite{xiao2025spatialtracker} & 0.188 & 0.094 & 2.487 \\
\rowcolor{GroupLeft}
+ BA w/o mask & 0.168 \textcolor{PineGreen}{(-0.020)} & 0.044 \textcolor{PineGreen}{(-0.050)} & 1.573 \textcolor{PineGreen}{(-0.914)} \\
\rowcolor{GroupLeft}
+ BA w/ our mask & 0.103 \textcolor{PineGreen}{(-0.085)} & 0.025 \textcolor{PineGreen}{(-0.069)} & 1.249 \textcolor{PineGreen}{(-1.238)} \\

\rowcolor{GroupRight}
VGGT~\cite{wang2025vggt} & 0.238 & 0.110 & 2.815 \\
\rowcolor{GroupRight}
+ BA w/o mask & 0.182 \textcolor{PineGreen}{(-0.056)} & 0.053 \textcolor{PineGreen}{(-0.057)} & 1.528 \textcolor{PineGreen}{(-1.287)} \\
\rowcolor{GroupRight}
+ BA w/ our mask & 0.178 \textcolor{PineGreen}{(-0.060)} & 0.051 \textcolor{PineGreen}{(-0.059)} & 1.530 \textcolor{PineGreen}{(-1.285)} \\

\rowcolor{GroupLeft}
Cut3r~\cite{Wang2024videocutler} & 0.260 & 0.057 & 1.189 \\
\rowcolor{GroupLeft}
+ BA w/o mask & 0.248 \textcolor{PineGreen}{(-0.012)} & 0.050 \textcolor{PineGreen}{(-0.007)} & 1.144 \textcolor{PineGreen}{(-0.045)} \\
\rowcolor{GroupLeft}
+ BA w/ our mask & 0.249 \textcolor{PineGreen}{(-0.011)} & 0.050 \textcolor{PineGreen}{(-0.007)} & 1.127 \textcolor{PineGreen}{(-0.062)} \\

\rowcolor{GroupRight}
TTT3r~\cite{chen2025ttt3r} & 0.227 & 0.060 & 1.437 \\
\rowcolor{GroupRight}
+ BA w/o mask & 0.178 \textcolor{PineGreen}{(-0.049)} & 0.037 \textcolor{PineGreen}{(-0.023)} & 1.431 \textcolor{PineGreen}{(-0.006)} \\
\rowcolor{GroupRight}
+ BA w/ our mask & 0.167 \textcolor{PineGreen}{(-0.060)} & 0.036 \textcolor{PineGreen}{(-0.024)} & 1.334 \textcolor{PineGreen}{(-0.103)} \\

\bottomrule
\end{tabular}
\label{tab:ba-right}
\end{minipage}
\end{table*}
\parahead{Implementation Details.} 
Following AnyCam~\cite{wimbauer2025anycam}, we use a two-stage refinement strategy to optimize the camera poses. The first stage employs a sliding-window optimization, where we use a window of eight frames, with a stride of one frame.  We choose a small stride to densely refine the initial pose estimation. 
Within each window, we build short-term correspondences by sampling a uniform 16x16 grid of points and tracking them across the subsequent 8 frames using predicted optical flow. 
For each window, we perform 400 optimization steps. 
In the second stage, we optimize all the camera poses and static points jointly 
for 5000 steps to ensure long-range coherence.

\parahead{Experimental Settings.} We evaluate our pose optimization pipeline with \ourmodel by comparing with two categories of state-of-the-art methods. Optimization-based methods include MegaSaM~\cite{zhang2024megasam}, AnyCam~\cite{wimbauer2025anycam}, 
and Monst3r~\cite{zhang2024monst3r}.
Feed-forword models include Easi3r~\cite{chen2025easi3r} SpatialTrackerV2~\cite{xiao2025spatialtracker}, VGGT~\cite{wang2025vggt}, Cut3r~\cite{wang2025continuous}, and TTT3r~\cite{chen2025ttt3r}. For all the methods, we take their camera pose estimation as initialization and optimize with the masks predicted from \ourmodel. For a fair comparison, we also perform the same pose optimization process and do not apply the motion filtering (without using the mask). For the baselines that also predict the motion masks, including MegaSaM~\cite{zhang2024megasam}, AnyCam~\cite{wimbauer2025anycam}, Monst3r~\cite{zhang2024monst3r}, and Easier~\cite{chen2025easi3r}, we perform additional experiments and optimize with their predicted motion masks for comparison.


\noindent{\bf Evaluation Datasets and Metrics.} We evaluate our method on two benchmark datasets: Sintel~\cite{Butler:ECCV:2012} and LightSpeed~\cite{rockwell2025dynamic}. 
During evaluation, we align the predicted trajectory with the ground-truth trajectory using Umeyama alignment~\cite{umeyama1991least} to handle the scale ambiguity of camera poses, following the practice. We compute three metrics: Absolute Trajectory Error (ATE) to measure the global consistency of the estimated trajectory, Relative Translation Error (RTE) to assess accuracy in translation, and Relative Rotation Error (RRE) to evaluate rotational.

\parahead{Experimental Results.}
We provide quantitative results in Table~\ref{tab:ba-left} for the challenging LightSpeed dataset. Results on Sintel are provided in the Appendix. 
By leveraging masks from \ourmodel to effectively remove dynamic objects from bundle adjustment, our approach enhances the accuracy of various state-of-the-art methods for camera pose estimation. The improvement is more significant on the challenging LightSpeed dataset, highlighting the efficacy of our method in dynamic environments.
When comparing using the different sources of motion masks, optimizing with the masks from our \ourmodel achieves consistently better performance compared to optimizing with the masks from each baseline, highlighting the benefits of obtaining precise motion masks. A qualitative example is provided in Fig~\ref{fig:pose_opt}.





\subsection{Depth Optimization}
\label{exp:depth_optimization}
\parahead{Implementation Details.} We follow the settings from MegaSAM~\cite{zhang2024megasam} to optimize video depth.
We set the weights for the optical flow, prior, and temporal consistency losses to $\lambda_{\mathsf{flow}}=1.0$, $\lambda_{\mathsf{prior}}=1.0$, and $\lambda_{\mathsf{temp}}=0.2$, respectively. The optimization is conducted at a resolution of $336 \times 144$, and the final depth maps are upsampled to $672 \times 288$ for evaluation.

\begin{figure}
    \vspace{-8mm}
    \centering
    \includegraphics[width=1\linewidth]{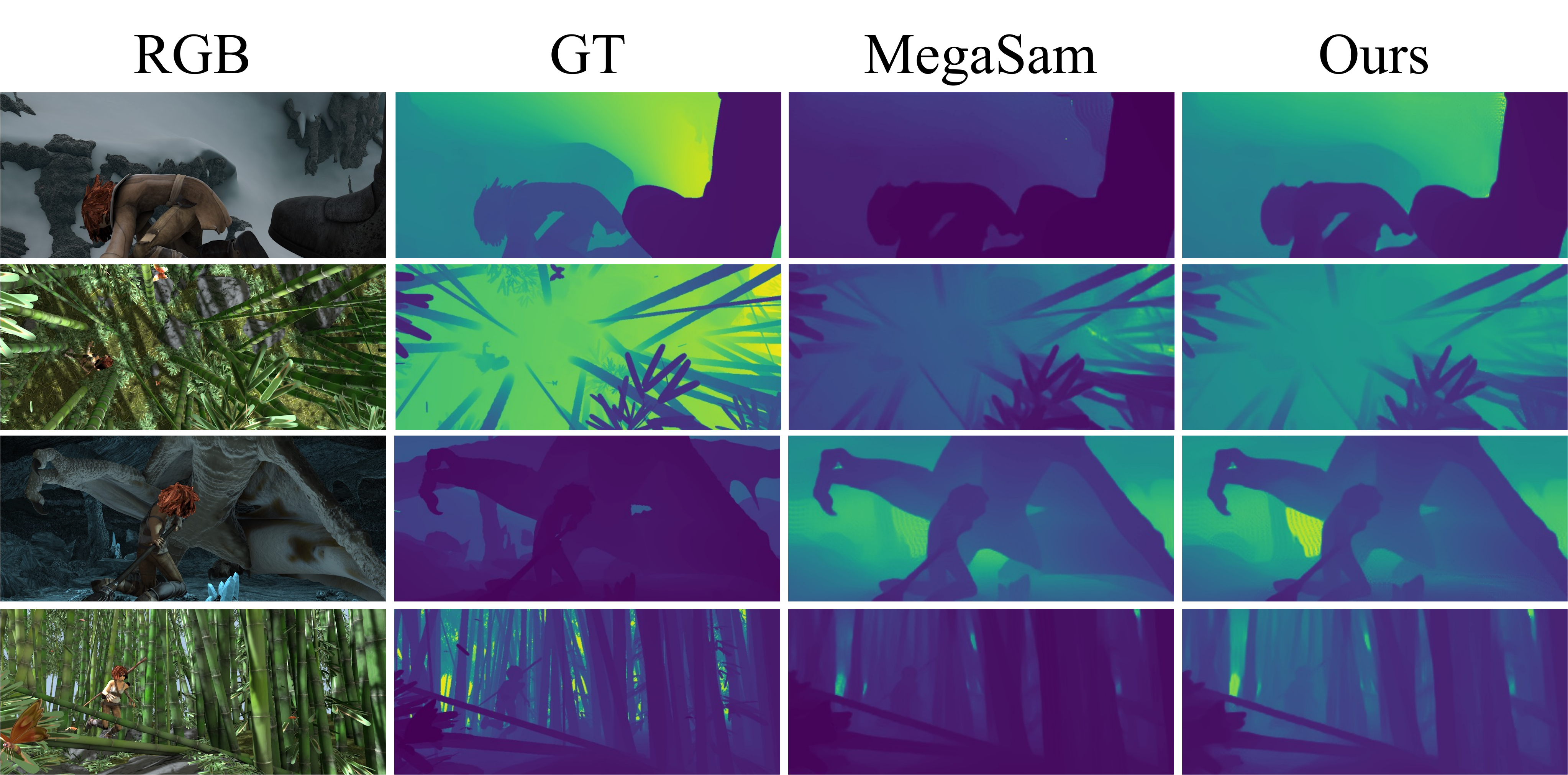}
    \vspace{-1em}
    \caption{Visual comparisons of depths. We compare with the original CVD~\cite{zhang2024megasam} and CVD with \ourmodel.}
    \label{fig:depth_cmp}
    \vspace{-2em}
\end{figure}

\parahead{Evaluation Datasets and Metrics.} We evaluate on three established benchmarks: MPI Sintel~\cite{Butler:ECCV:2012}, DyCheck~\cite{gao2022monocular}, and TUM-RGBD~\cite{sturm12iros}.
For Sintel, following prior work~\cite{zhang2022structure}, our evaluation is performed on the 18 sequences. 
For DyCheck, 
we use the ground truth camera poses and depth maps provided by Shape of Motion~\cite{wang2025shape}. 
We compute standard depth estimation metrics for evaluation: absolute relative error (abs-rel), log root-mean-square error (log-rmse), and the threshold accuracy ($\delta_{1.25}$).

\parahead{Experimental Settings.} We compare with MegaSaM~\cite{zhang2024megasam} by applying the same consistent video depth optimization process and only changing the source of the masks. We use the mask either from the original MegaSaM or use the mask predicted from \ourmodel.

\parahead{Experimental Results.} Table~\ref{table:sintel_depth} shows the quantitative results, and Fig.~\ref{fig:depth_cmp} provides qualitative comparison. 
Our method consistently improves depth accuracy across all datasets. 
This performance gain is attributed to our discrete, semantics-aware motion mask, which replaces the learned, ambiguous uncertainty map used in the original CVD formulation. \ourmodel provides a clear signal that isolates static regions for the optimization, leading to more robust and accurate depth.

\begin{table}[t]
\begin{center}
\scriptsize
\caption{Quantitative comparisons of video depths. Lower is better for abs-rel and log-rmse, and higher is better for $\delta_{1.25}$.} 
\definecolor{GroupLeft}{RGB}{240,248,255}
\definecolor{GroupRight}{RGB}{255,240,245}
\begin{tabular}{l ccc}
\toprule
& \multicolumn{3}{c}{Sintel~\cite{Butler:ECCV:2012}} \\
Method & abs-rel $\downarrow$ & log-rmse $\downarrow$ & $\delta_{1.25}$ $\uparrow$ \\
\midrule
\rowcolor{GroupLeft}
MegaSam~\cite{zhang2024megasam} & 0.3081 & 0.4755 & 59.30 \\
\rowcolor{GroupLeft}
+ CVD w/ their mask & 0.2369 & 0.4161 & 71.31 \\
\rowcolor{GroupLeft}
+ CVD w/ our mask & 0.2135 & 0.3957 & 72.77 \\



\bottomrule
\midrule
& \multicolumn{3}{c}{Dycheck~\cite{gao2022monocular}} \\
Method & abs-rel $\downarrow$ & log-rmse $\downarrow$ & $\delta_{1.25}$ $\uparrow$ \\
\midrule
\rowcolor{GroupLeft}
MegaSam~\cite{zhang2024megasam} & 0.1411 & 0.2158 & 91.61 \\
\rowcolor{GroupLeft}
+ CVD w/ their mask & 0.1083 & 0.2006 & 94.11 \\
\rowcolor{GroupLeft}
+ CVD w/ our mask & 0.1069  & 0.1998 & 94.53 \\
\bottomrule
\midrule
& \multicolumn{3}{c}{TUM-RGBD~\cite{sturm12iros}} \\
Method & abs-rel $\downarrow$ & log-rmse $\downarrow$ & $\delta_{1.25}$ $\uparrow$ \\
\midrule
\rowcolor{GroupLeft}
MegaSam~\cite{zhang2024megasam} & 0.1777 & 0.2595 & 93.63 \\
\rowcolor{GroupLeft}
+ CVD w/ their mask & 0.1598 & 0.2523 & 95.13 \\
\rowcolor{GroupLeft}
+ CVD w/ our mask & 0.1574  & 0.2519  & 95.17 \\
\bottomrule

\end{tabular}
\vspace{-0.5em}
\label{table:sintel_depth} 
\vspace{-5mm}
\end{center} 
\end{table}

\subsection{4D Track Optimization}
\label{exp:track_optimization}
\vspace{-1mm}

\parahead{Experimental Settings.}
We follow the same experimental settings from Stereo4D~\cite{jin2025stereo4d} for evaluation, and only replace the motion score from the original paper with our dynamic mask, as mentioned in the Sec.~\ref{subsec:track_optimization}. We test on the real-world videos released by the Stereo4D paper. The hyperparameters and initial camera parameters, depth, and optical flow are the same as Stereo4D.

\begin{figure}[t]
    \vspace{-2mm}
    \centering
    \includegraphics[width=1\linewidth]{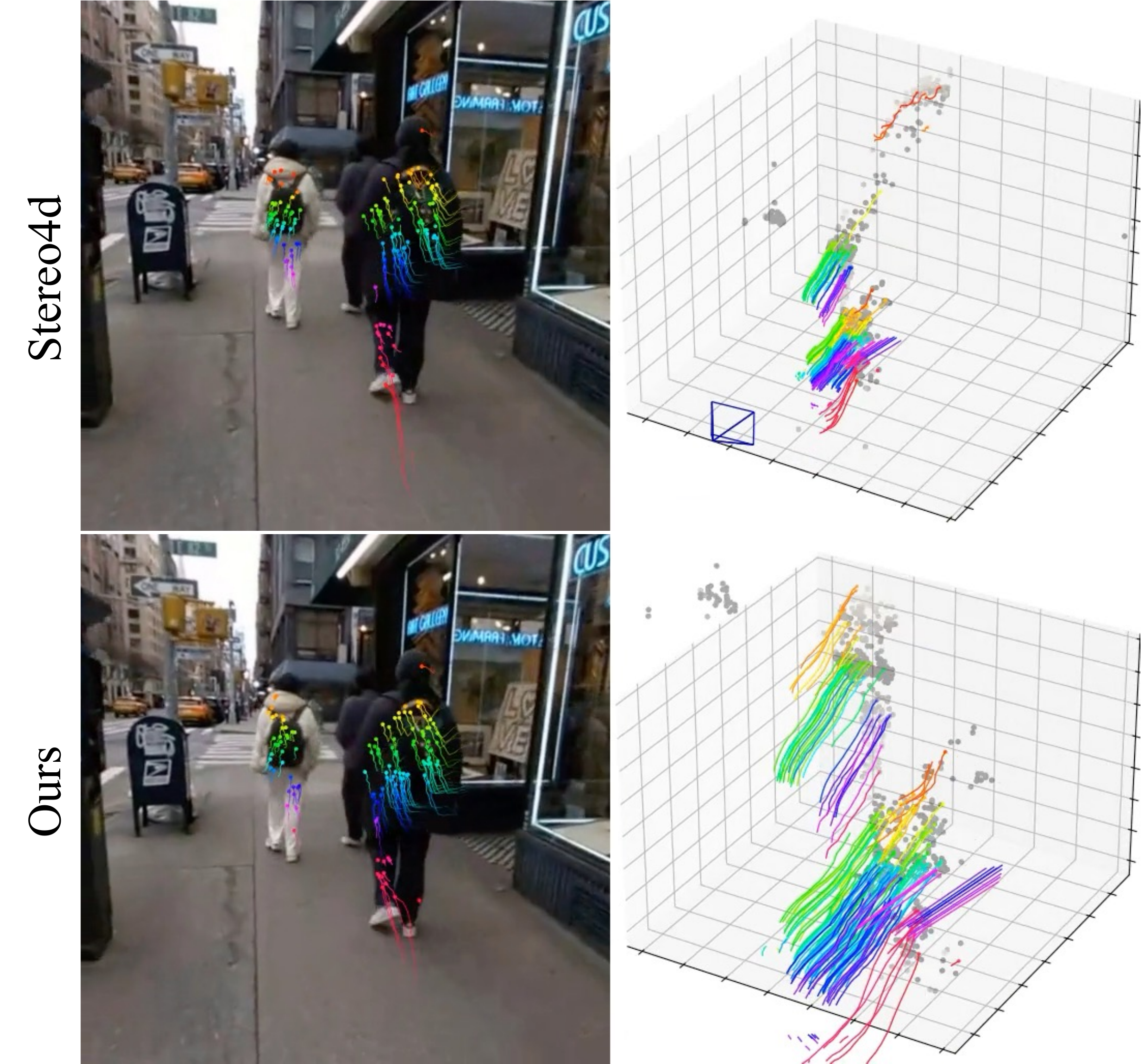}
    \vspace{-6mm}
    \caption{Comparison of the optimized tracks. Using the dynamic mask from \ourmodel, we can generate more plausible and coherent motion for the dynamic object.} 
    \label{fig:tracks_cmp}
    \vspace{-5mm}
\end{figure}

\parahead{Experimental Results.}
We provide qualitative comparisons with Stereo4D in Fig.~\ref{fig:tracks_cmp}. Our 4D track optimization produces more plausible trajectories on dynamic objects compared to the original Stereo4D~\cite{jin2025stereo4d}. The trajectories refined by our method exhibit more physically plausible motion and maintain a more coherent relative structure for dynamic objects. The key to this improvement is our dynamic object masks. The motion score of Stereo4D can be noisy, leading to ambiguities in balancing the static and dynamic loss. \ourmodel can well disentangle the dynamic and static objects, allowing the optimizer to confidently apply the appropriate static or dynamic regularization to the corresponding regions.

\section{Conclusion}
\vspace{-2mm}
\parahead{Conclusion.} In this work, we presented the Dynamic Prior (\ourmodel), a generalizable motion segmentation framework that leverages the reasoning capabilities of Vision Language Models and the segmentation capacity of SAM2. 
By providing robust and accurate dynamic masks without task-specific training, \ourmodel facilitates broad tasks for robust and accurate 3D structure understanding from casual real-world videos. It enables more reliable camera pose estimation, depth reconstruction, and 4D trajectory recovery. 
Extensive experiments on both synthetic and real-world videos demonstrate that \ourmodel not only achieves state-of-the-art performance on motion segmentation, but also significantly improves accuracy for 3D scene understanding. 

\parahead{Limitations.}
While our method effectively improves the performance of camera pose, depth, and tracking optimization, these three tasks are independently performed. Studying a cohesive framework for joint optimization is an exciting future work for research. Furthermore, since our dynamic mask is only used to filter out dynamic objects, it will fail in scenarios where a large dynamic object dominates the field of view. Our motion segmentation pipeline relies on text to prompt SAM2.  While effective in many cases, this language-based interface can be a source of ambiguity when multiple objects of the same semantic category are present (e.g., a group of animals). Differentiating between these instances using purely textual prompts can be challenging, potentially leading to incorrect segmentation. Exploring an alternative, implicit feature-level communication between the reasoning and segmentation model could address this limitation and is a promising avenue for future research.


{
    \small
    \bibliographystyle{ieeenat_fullname}
    \bibliography{main}
}

\clearpage
\setcounter{page}{1}
\maketitlesupplementary
\appendix
\section{Dynamic Object Reasoning Prompt}
\label{sec:dsua_cot}

\begin{figure*}
\vspace{-8mm}
\begin{minipage}[h]{0.25\textwidth}

\centering\raisebox{\dimexpr \topskip-\height}{%
  \includegraphics[width=\linewidth]{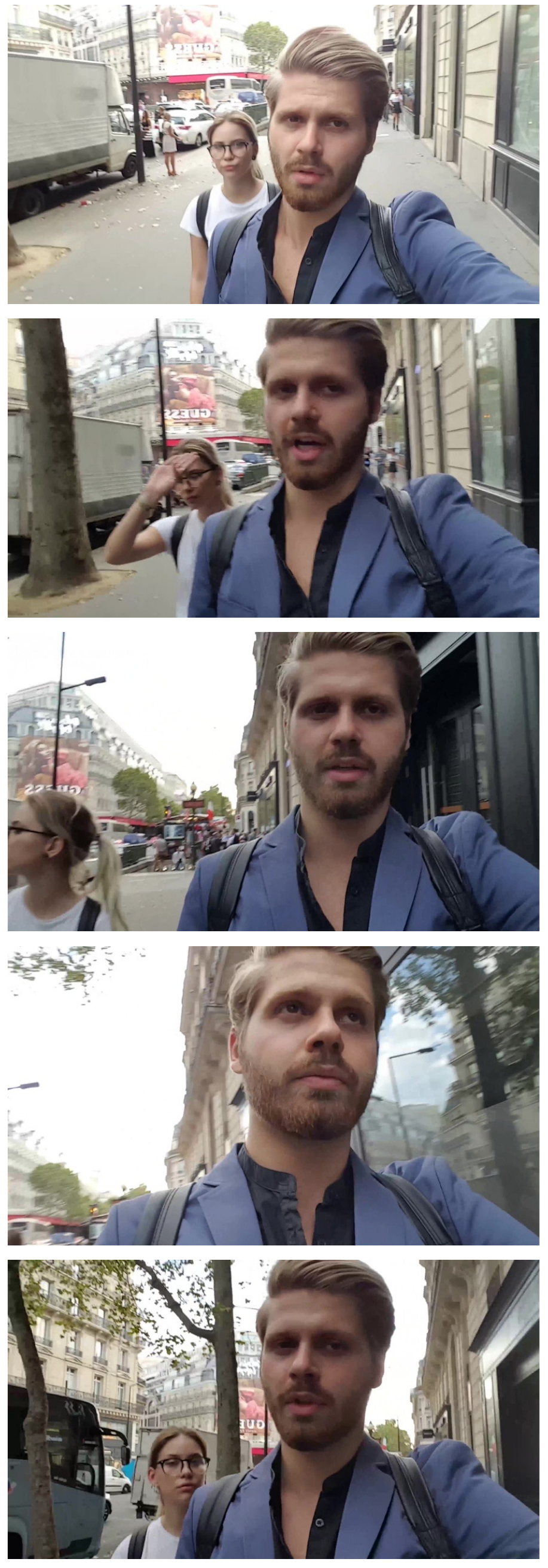}}
  \captionof{figure}{Sampled Frames.}
  \label{fig:CoT}
\end{minipage}\hfill
\begin{minipage}[h]{0.72\textwidth}
    \par\noindent\rule{\linewidth}{0.3pt}
    \textbf{An example of dynamic object reasoning (GPT-4o):}
    \vspace{-0.5em}
    \par\noindent\rule{\linewidth}{0.3pt}
    \noindent{\small\textbf{Chain of Thoughts:}} 
    \small{
    \noindent{\small\textit{- Frame 1: }}
  - *What can be seen in the frame?* A man and a woman walking on a sidewalk. There are vehicles and buildings in the background.  
  - *How many objects are moving?* Two main moving objects: the man and the woman.  
  - *What is the relationship between these objects?* They appear to be walking together.
  - *Are any objects obscured or camouflaged?* No significant obstructions; both individuals are clearly visible.
  - *Is this a good keyframe for any object?* Yes, both individuals are clearly visible.

    \noindent{\small\textit{- Frame 2: }}
    - What can be seen in the frame?* The same man and woman, with the woman slightly obscured by the man.  
    - *How many objects are moving?* Two main moving objects: the man and the woman.  
    - *What is the relationship between these objects?* They are still walking together.  
    - *Are any objects obscured or camouflaged?* The woman is partially obscured by the man.  
    - *Is this a good keyframe for any object?* No, the woman is partially obscured.

    \noindent{\small\textit{- Frame 3: }}
    - *What can be seen in the frame?* The man is in the foreground, and the woman is turning away.  - *How many objects are moving?* Two main moving objects: the man and the woman.  - *What is the relationship between these objects?* They are still walking together, but the woman is turning.  - *Are any objects obscured or camouflaged?* The woman is turning away, making her less visible.  - *Is this a good keyframe for any object?* No, the woman is not clearly visible.-

    \noindent{\small\textit{- Frame 4: }}
    - *What can be seen in the frame?* The man is in the foreground, and the woman is not visible.  - *How many objects are moving?* Only the man is visible and moving.  - *What is the relationship between these objects?* The woman is not visible, so only the man is considered.  - *Are any objects obscured or camouflaged?* The woman is completely out of frame.  - *Is this a good keyframe for any object?* No, the woman is not visible.
    
    \noindent{\small\textit{- Frame 5: }}
    - *What can be seen in the frame?* The man and woman are both visible again, walking together.  - *How many objects are moving?* Two main moving objects: the man and the woman.  - *What is the relationship between these objects?* They are walking together.  - *Are any objects obscured or camouflaged?* No significant obstructions; both individuals are clearly visible.  - *Is this a good keyframe for any object?* Yes, both individuals are clearly visible.
   
}

\noindent{\small\textbf{\textit{Output List:}}} 
[\{object\_index: 1, keyframe: 1, object\_description: \"the man in the blue jacket\"\}, \{object\_index: 2, keyframe: 1, object\_description: \"the woman with glasses\"\}, \{object\_index: 1, keyframe: 5, object\_description: \"the man in the blue jacket\"\}, \{object\_index: 2, keyframe: 5, object\_description: \"the woman with glasses\"\}]
\vspace{-0.5em}
    \par\noindent\rule{\linewidth}{0.3pt}
\end{minipage}
\end{figure*}

One of the key components of our framework is the dynamic object reasoning, which leverages a chain-of-thought process to perform temporal-semantic analysis. To prompt the MLLM, we first subsample the keyframe candidates $\{\keycand_i\}_{i=1}^{T'}$, then we use the following text prompt:
\begin{adjustwidth}{2em}{2em}
\par\noindent\rule{\linewidth}{0.3pt}
\textbf{Chain-of-Thought Prompt for Dynamic Object Reasoning:}
\vspace{-1em}
\par\noindent\rule{\linewidth}{0.3pt}
\noindent{\small\textit{`You will act as a keyframe selection agent for a video reasoning task. During each inference, you will be given multiple keyframes sampled from a long video. The keyframes are presented as separate images in strict temporal order (Frame 1, Frame 2, ...). You need to find all moving objects in the given video. You need to think in chain of thoughts to analyze each keyframe and find the best keyframes for each target object, where a segmentation model can find the target object in that frame with less effort. Your chain of thoughts should begin with what can be seen in each keyframe, how many objects in total, etc. Some of the objects may be seriously obscured or blocked by other objects. Some of the objects may be camouflaged in their surroundings. Analyze each frame separately to get all the visible objects. This chain of thoughts should follow the output format:}}

\noindent{\small\textit{``Chain of Thoughts:}}

\noindent\small\textit{- Frame 1: \scless analysis of frame 1\scgreater;}

\noindent\small\textit{- Frame 2: \scless analysis of frame 2\scgreater;}

\noindent{\small\textit{...'', where you have to ask questions to yourself and answer them. Your answer should be as detailed as possible. You should start with broader questions, like ``what can be seen in the frame?'' to some detailed questions like ``how many and which objects are moving?'', ``What is the relationship between these objects?'', etc. There will be many questions and answers in the analysis of each frame, helping you to fully understand the frame. The actual questions vary by cases. Generate the questions and answers based on the analysis of each frame. Additionally, determine whether any target object becomes fully occluded or goes out of frame and later reappears. Count the number of continuous visibility periods for each object (from first appearance to disappearance/end, then from reappearance to next disappearance/end). Your thinking process aims to find the keyframe for each target object of interest (find all the target objects, each of which may correspond to one keyframe per continuous visibility period). Lastly, you have to output a list of dictionaries with the format:}}

\noindent{\small\textit{``Output list: [\{object\_index: 1, keyframe: k\_1, object\_description: \scless description of the object 1 in keyframe k\_1\scgreater\}, \{object\_index: 2, keyframe: k\_2, object\_description: \scless description of the object 2 in keyframe k\_2\scgreater\}, ...]''}}

\noindent{\small\textit{, where each element in the list is a dictionary with three items: object index, keyframe index, object description. k is the k-th keyframe in the sequence (1-based index), counted by temporal order from Frame 1 to Frame N. object\_index is a numbering integer starting with 1. object\_description implies the description for that object in a particular frame, helping the model to find the object in that particular frame. For example, a valid element in an output list can be like 'Output list: [\{object\_index: 1, keyframe: 4, object\_description: ``the man at the top left corner of the image''\}]'. In case there are multiple objects in the same keyframe, you need to discriniate and describe them separetly. DO NOT describe multiple objects in the same keyframe as a single object. Include the objects even if they are only partially visible. If an object disappears (fully occluded or out of frame) and later reappears, treat each continuous visibility span as a separate period and select one best keyframe per period. Reuse the same object\_index for all periods of the same object and output multiple entries—one per period. Consider the period in your question when analyzing the frame. While choosing the keyframe for any object, you should prioritize those frames where objects are not overlapped. This will help the model to better recognize the object. Keep the output list in text format. Don't use JSON formatting. The output list begins with the prefix ``Output list: '', followed by a square bracket with multiple curly brackets. The square bracket should be in the same line, following the format ``Output list: [...]''. Don't start with a new line.}}

\small\textit{Here is a sequence with \{num\_keyframes\} keyframes presented as separate images in temporal order. Follow the instruction and output the index of the best keyframe.'}
\vspace{-1em}
\par\noindent\rule{\linewidth}{0.3pt}
\end{adjustwidth}

where \texttt{\{num\_keyframes\}} are placeholders that are dynamically filled based on the input. The MLLM's output is then parsed to extract the selected keyframes and object descriptions. Fig~\ref{fig:CoT} provides an example of the reasoning process and the outputs.

\section{Depth Optimization}
\label{appendix_subsec:depth_optimization}

The optimization function is defined by Eq.~\ref{eq:depth_loss}, and we introduce $\mathcal{L}_{\text{flow}}$ in Eq.~\ref{eq:flow}. Here we give definitions of $\mathcal{L}_{\text{temp}}$ and $\mathcal{L}_{\text{prior}}$.
The temporal consistency loss, $\mathcal{L}_{\text{temp}}$, ensures smooth depth evolution. It matches the warped depth from frame $i$ to $j$, $d_{i \to j}(\mathbf{p}) = (\mathbf{P}^{i \to j} \pi_K^{-1}(\mathbf{p}, \hat{\mathbf{D}}_i(\mathbf{p})))_z$, with the depth at the flow-warped pixel in frame $j$, $\hat{\mathbf{D}}_j(\mathbf{p} + \mathbf{F}^{i \to j}(\mathbf{p}))$, where $\hat{\mathbf{D}}$ is the optimized depth. The loss is:
\begin{equation}
    \mathcal{L}_{\text{temp}}^{i \to j} = \bm{\Omega}_i(\mathbf{p}) \cdot \delta(d_{i \to j}(\mathbf{p}), \hat{\mathbf{D}}_j(\mathbf{p} + \mathbf{F}^{i \to j}(\mathbf{p}))) + \log(1/\bm{\Omega}_i(\mathbf{p})),
    \label{eq:temp}
\end{equation}
where $\delta(a,b) = ||\max(a/b, b/a)||_1$.

The mono-depth prior loss, $\mathcal{L}_{\text{prior}}$, regularizes the optimized depth to remain faithful to the initial estimates. It consists of three components:
\begin{equation}
    \mathcal{L}_{\mathsf{prior}} = \mathcal{L}_{\mathsf{si}} + \lambda_{\mathsf{grad}} \mathcal{L}_{\mathsf{grad}} + \lambda_{\mathsf{normal}} \mathcal{L}_{\mathsf{normal}},
\end{equation}
where $\mathcal{L}_{\mathsf{si}}$ is a scale-invariant loss, $\mathcal{L}_{\mathsf{grad}}$ is a gradient consistency loss, and $\mathcal{L}_{\mathsf{normal}}$ is a surface normal loss. Let $R_t(\mathbf{p}) = \log(\hat{\mathbf{D}}_t(\mathbf{p})) - \log(\mathbf{D}_t(\mathbf{p}))$ be the logarithmic depth difference, where $\mathbf{D}_t$ is the initial monocular depth. The scale-invariant loss is:
\begin{equation}
    \mathcal{L}_{\mathsf{si}} = \frac{1}{N_p} \sum_{\mathbf{p}} (R_t(\mathbf{p}))^2 - \frac{1}{N_p^2} \left( \sum_{\mathbf{p}} R_t(\mathbf{p}) \right)^2,
\end{equation}
where $N_p$ is the number of pixels. The gradient consistency loss penalizes disparities in the log-depth map gradients across multiple scales:
\begin{equation}
    \mathcal{L}_{\mathsf{grad}} = \frac{1}{N_p} \sum_s \sum_{\mathbf{p}} w_{\nabla}^{s}(\mathbf{p}) \left( |\nabla_x R_t^s (\mathbf{p})| + |\nabla_y R_t^s(\mathbf{p})| \right),
\end{equation}
where $R_t^s$ is the logarithmic depth difference on scale $s$, $w_{\nabla}^{s}(\mathbf{p})$ is a weighting term, and $\nabla_x$ and $\nabla_y$ are the gradients of the image along the x and y directions. The surface normal loss preserves geometry:
\begin{equation}
    \mathcal{L}_{\mathsf{normal}} = \sum_{\mathbf{p}} (1 - \hat{\mathbf{N}}_t(\mathbf{p}) \cdot \mathbf{N}_t(\mathbf{p})),
\end{equation}
where $\hat{\mathbf{N}}_t$ and $\mathbf{N}_t$ are surface normals from the optimized and prior depth maps. We set $\lambda_{\mathsf{grad}}=1$ and $\lambda_{\mathsf{normal}}=4$.

\section{4D Track Optimization}

The optimization objective for 4D track optimization is defined in Eq.~\ref{eq:stereo4d}. Here we provide the detailed definitions for the three loss terms: $\mathcal{L}_{\mathsf{static}}$, $\mathcal{L}_{\mathsf{dynamic}}$, and $\mathcal{L}_{\mathsf{reg}}$.

The static loss $\mathcal{L}_{\mathsf{static}}$ encourages stationary points to maintain a consistent position in world space:
\begin{equation}
\mathcal{L}_{\mathsf{static}} = \sum_{i=1}^{N} \sum_{j=1}^{N} \frac{\| \mathbf{x}_i' - \mathbf{x}_j' \|^2}{N^2}.
\end{equation}
The dynamic loss $\mathcal{L}_{\mathsf{dynamic}}$ promotes smoother trajectories for moving objects by minimizing acceleration along the camera ray. The acceleration is computed over multiple window sizes $\mathcal{W}=\{1,3,5\}$:
\begin{equation}
\mathcal{L}_{\mathsf{dynamic}} = \sum_{i=1}^{N} \sum_{\Delta\in\mathcal{W}} \left[ \left( \mathbf{x}_{i+\Delta}' - 2\mathbf{x}_i' + \mathbf{x}_{i-\Delta}' \right)^\top \mathbf{r}_i \right]^2.
\end{equation}
Finally, the regularization loss $\mathcal{L}_{\mathsf{reg}}$ ensures that the refined tracks remain faithful to the original depth information:
\begin{equation}
\mathcal{L}_{\mathsf{reg}} = \lambda_{\mathsf{reg}} \sum_{i=1}^{T} \left( \frac{1}{\delta_i + \|\mathbf{x}_i-\mathbf{c}_i\|} - \frac{1}{\|\mathbf{x}_i-\mathbf{c}_i\|} \right)^2.
\end{equation} 
We set $\lambda_{\mathsf{reg}}=0.1$. The camera center $\mathbf{c}_i$ is the translation component of the camera-to-world pose $\mathbf{P}_i$, which can be formulated as a $4\times 4$ matrix:
\begin{equation}
    \mathbf{P}_i = 
\begin{pmatrix} 
\mathbf{R}_i & \mathbf{c}_i \\
\mathbf{0}^\top & 1 
\end{pmatrix},
\end{equation}
where $\mathbf{R}_i$ is the rotation matrix.

\section{More Analysis for Motion Segmentation}
In this part, we give more analysis of dynamic object reasoning and dynamic object segmentation. All ablation studies are applied on FBMS-59~\cite{ochs2014fbms59} and we employ two standard metrics. 
The region similarity ($\mathcal{J}$), also known as the Jaccard index, measures the overlap between the predicted and ground-truth masks.
The contour similarity ($\mathcal{F}$) measures the accuracy of the predicted object boundaries by computing a score based on precision and recall.

\subsection{Analysis of Dynamic Object Reasoning}
We perform an analysis of various Vision Language Models (VLMs) for dynamic object reasoning, with the results detailed in Table~\ref{tab:dsua}. Among closed-source models, GPT-4o~\cite{hurst2024gpt} achieves the highest performance, leading us to select it as our default dynamic object reasoning. We also evaluated several open-source alternatives, including Qwen2.5-VL~\cite{bai2025qwen2}, Qwen3-VL~\cite{yang2025qwen3}, and InternVL3.5~\cite{wang2025internvl3}. The performance of these models generally aligns with the scaling law, where larger models tend to yield better results.

\begin{table*}[!t]
\vspace{-4mm}
\centering
\small
\setlength{\tabcolsep}{3.5pt}
\renewcommand{\arraystretch}{1} 

\begin{minipage}[t]{0.35\textwidth}
\centering
\caption{Quantitative comparison with different VLMs for dynamic object reasoning on FBMS-59 testset.}
\begin{tabular}{@{}lccc@{}}
\toprule
\multirow{2}{*}{Methods} & \multicolumn{3}{c}{FBMS-59} \\ 
\cmidrule(l){2-4} 
& $\mathcal{J\&F} \uparrow$ & $\mathcal{J} \uparrow$ & $\mathcal{F}\uparrow$ \\ 
\midrule
GPT-4o~\cite{hurst2024gpt} & 86.47 & 83.99 & 88.95 \\
Gemini2.5-Pro~\cite{comanici2025gemini} & 85.86 & 83.45 & 88.26 \\
\midrule
Qwen2.5-VL-7B~\cite{bai2025qwen2} & 82.27 & 79.79 & 84.75 \\
Qwen2.5-VL-32B~\cite{bai2025qwen2} & 83.80 & 80.79 & 86.80 \\
Qwen3-VL-8B~\cite{yang2025qwen3} & 78.82 & 76.02 & 81.61 \\
InternVL3.5-8B~\cite{wang2025internvl3} & 78.36 & 75.08 & 81.63 \\
InternVL3.5-38B~\cite{wang2025internvl3} & 83.31 & 80.53 & 86.09 \\
\bottomrule
\end{tabular}
\label{tab:dsua}
\end{minipage}
\hfill
\begin{minipage}[t]{0.32\textwidth}
\centering
\caption{Quantitative comparison for dynamic object segmentation with different MLLM backbones on FBMS-59 testset.}
\begin{tabular}{@{}lccc@{}}
\toprule
Base MLLM & $\mathcal{J\&F} \uparrow$ & $\mathcal{J} \uparrow$ & $\mathcal{F} \uparrow$ \\
\midrule
InternVL2.5-1B~\cite{chen2024expanding} & 83.40 & 80.39 & 86.42 \\
InternVL2.5-4B~\cite{chen2024expanding} & 87.71 & 85.23 & 90.19 \\
InternVL2.5-8B~\cite{chen2024expanding} & 86.47 & 83.99 & 88.95 \\
InternVL2.5-26B~\cite{chen2024expanding} & 83.04 & 80.05 & 86.03 \\
\midrule
InternVL3-2B~\cite{zhu2025internvl3} & 83.60 & 80.73 & 86.47  \\
InternVL3-8B~\cite{zhu2025internvl3} & 81.46 & 78.22 & 84.69  \\
InternVL3-14B~\cite{zhu2025internvl3} & 81.22  & 77.84 & 84.60  \\
\midrule
Qwen2.5-VL-3B~\cite{bai2025qwen2} & 83.01 & 80.81 & 85.21  \\
Qwen2.5-VL-7B~\cite{bai2025qwen2} & 82.48 & 79.71 & 85.26  \\
\midrule
Qwen3-VL-4B~\cite{yang2025qwen3} & 81.90 & 78.94 & 84.86  \\
\bottomrule
\end{tabular}
\label{tab:MLLM}
\end{minipage}
\hfill
\begin{minipage}[t]{0.32\textwidth}
\centering
\caption{Quantitative comparison with different sampling strategies for dynamic object reasoning on FBMS-59 testset.}
\begin{tabular}{@{}lccc@{}}
\toprule
\multirow{2}{*}{Methods} & \multicolumn{3}{c}{FBMS-59} \\ 
\cmidrule(l){2-4} 
& $\mathcal{J\&F} \uparrow$ & $\mathcal{J} \uparrow$ & $\mathcal{F}\uparrow$ \\ 
\midrule
$|\keycand|=8$ & \multicolumn{3}{c}{Exceed Tokens} \\
$|\keycand|=16$ & 86.47 & 83.99 & 88.95 \\
$|\keycand|=32$ & 82.35 & 79.08 & 85.61 \\
$|\keycand|=64$ & \multicolumn{3}{c}{Fail} \\
\bottomrule
\end{tabular}
\label{tab:frame_sampling}
\end{minipage}
\end{table*}

\subsection{Analysis of MLLM in Dynamic Object Segmentation.}
\begin{figure}
    \centering
    \includegraphics[width=1\linewidth]{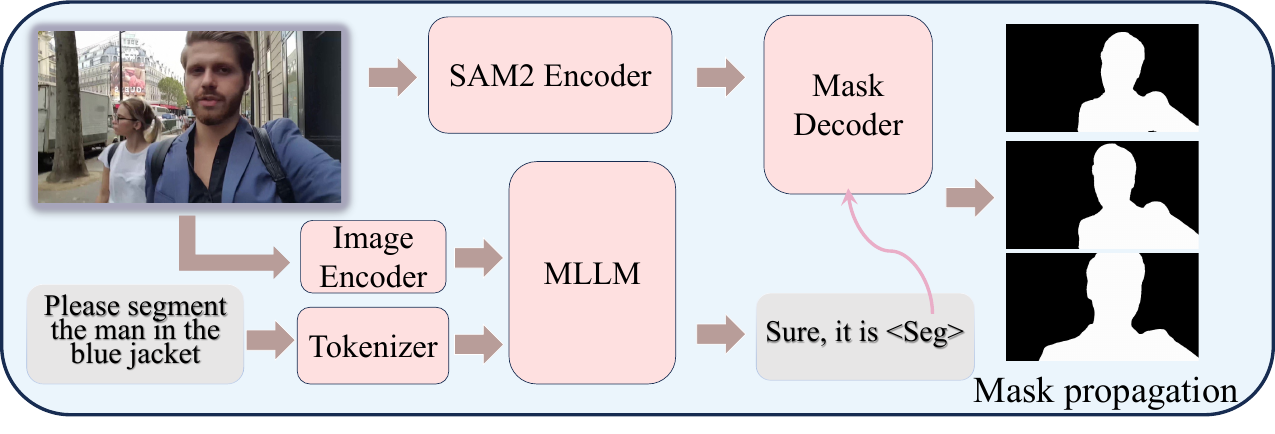}
    \caption{Details of dynamic object segmentation. The input image and text are encoded to an aligned token embedding space and fed to an MLLM. The instruction token ([Seg]) is the prompt for the mask decoder to generate an instance-level mask.}
    \label{fig:reasoning_segmentation}
    \vspace{-6mm}
\end{figure}

Fig.~\ref{fig:reasoning_segmentation} shows the details of workflow of dynamic object segmentation. Table.~\ref{tab:MLLM} presents a comprehensive overview of dynamic object segmentation based on recent MLLMs, including Intern2.5-VL~\cite{cai2024internlm2}, Intern3-VL~\cite{zhu2025internvl3}, Qwen2.5-VL~\cite{bai2025qwen2} and Qwen3-VL~\cite{yang2025qwen3}. 

\subsection{Analysis of video frames sampling.}
We investigate the impact of the $\keycand$, with results presented in Table~\ref{tab:frame_sampling}. The sampling process, as detailed in Section~\ref{sec:dynamic_obj_reason}, involves evenly selecting frames from the video. Our findings reveal a trade-off between sampling density and computational constraints. For a dense sampling rate ($|\keycand|=8$), the number of visual tokens exceeds the API rate limits of GPT-4o, preventing us from obtaining a result. Conversely, performance degrades as the sampling becomes sparser. At $|\keycand|=16$, the model achieves optimal performance, effectively capturing the scene dynamics. However, when the sampling interval increases to $|\keycand|=32$, we observe a noticeable drop in accuracy. At an even sparser rate of $|\keycand|=64$, the temporal information becomes insufficient for the VLM to reliably understand the dynamic evolution of the scene, leading to a failure in the analysis. This experiment underscores the importance of selecting an appropriate sampling rate that balances rich visual context with practical system limitations.

\section{More Motion Segmentation Results}
\begin{figure}
    \centering
    \includegraphics[width=1\linewidth]{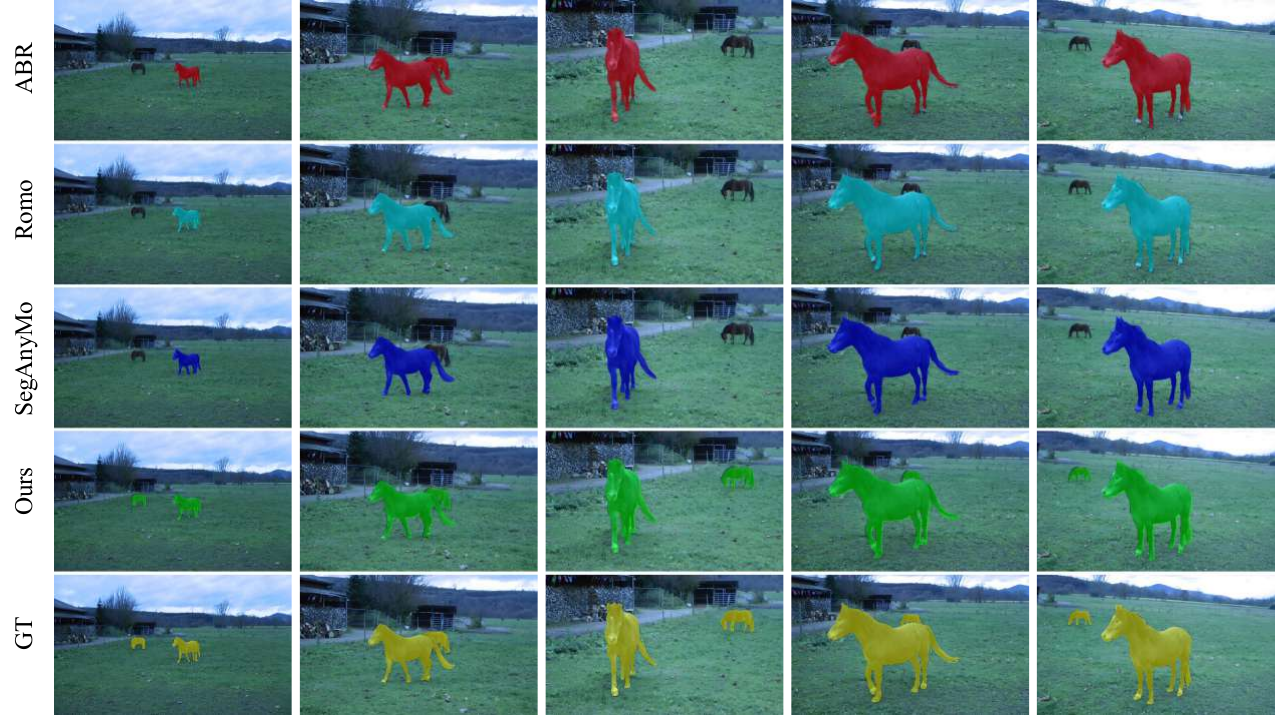}
    \caption{Visual comparison of moving object segmentation on FBMS-59~\cite{ochs2014fbms59} horses02.}
    \label{fig:fbms_result_appendix_1_compressed}
    \vspace{-6mm}
\end{figure}

\begin{figure}
    \centering
    \includegraphics[width=1\linewidth]{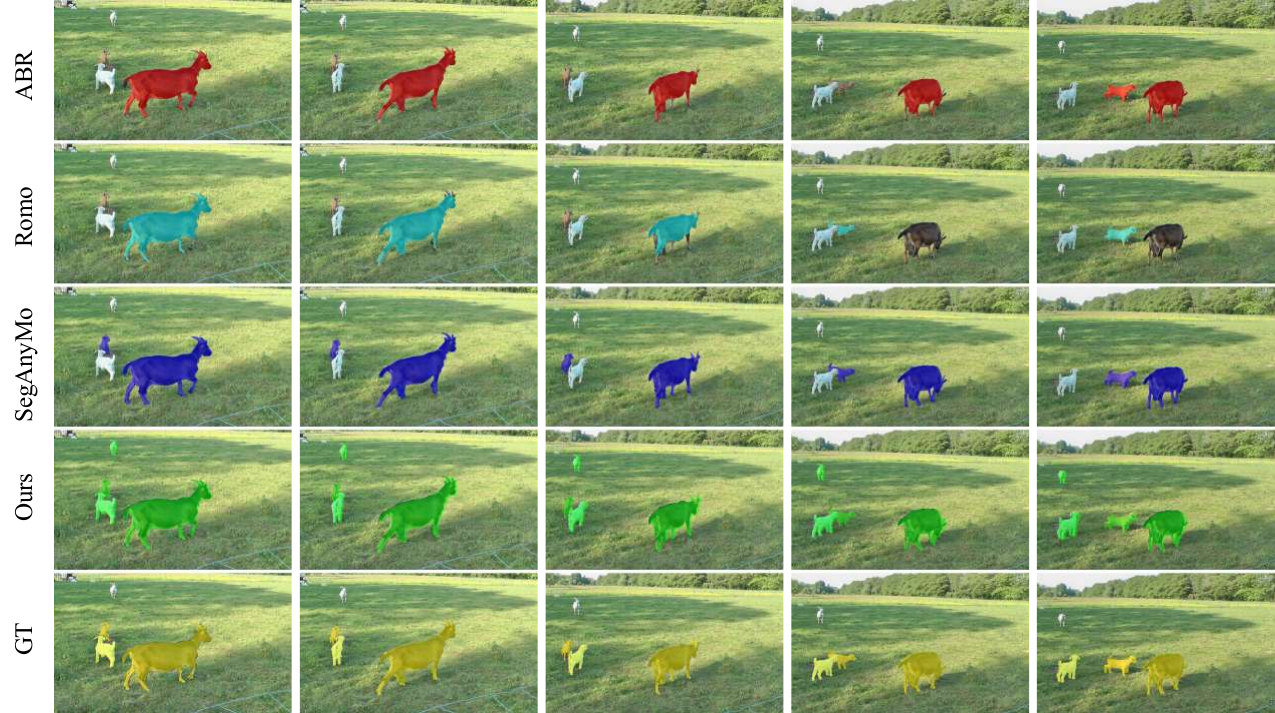}
    \caption{Visual comparison of moving object segmentation on FBMS-59~\cite{ochs2014fbms59} goats01.}
    \label{fig:fbms_result_appendix_2_compressed}
    \vspace{-6mm}
\end{figure}

Fig.~\ref{fig:fbms_result_appendix_1_compressed} and Fig.~\ref{fig:fbms_result_appendix_2_compressed} show more motion segmentation comparisons. \ourmodel has a better capability of dynamic scene reasoning, generating better dynamic segmentation masks compared to our baselines.

\section{More Pose Optimization Results}
\begin{table*}[!t]
\vspace{-2mm}
\label{tab:ba_sintel}
\centering
\scriptsize
\caption{Quantitative comparison for camera pose estimation with different baselines on Sintel~\cite{Butler:ECCV:2012} dataset.
}
\setlength{\tabcolsep}{5pt}
\renewcommand{\arraystretch}{1.2}
\definecolor{GroupLeft}{RGB}{240,248,255}
\definecolor{GroupRight}{RGB}{255,240,245}

\begin{minipage}[t]{0.4\textwidth} 
\centering
\begin{tabular}{@{}l c c c@{}}
\toprule
\multicolumn{4}{@{}c}{\textbf{Methods that predict motion masks}} \\
\cmidrule(lr){1-4}
Method & ATE $\downarrow$ & RTE $\downarrow$ & RRE $\downarrow$ \\
\midrule

\rowcolor{GroupLeft}
MegaSam~\cite{zhang2024megasam} & 0.096 & 0.014  & 0.041 \\
\rowcolor{GroupLeft}
+ BA w/o mask & 0.141 \textcolor{red}{(+0.045)} & 0.038 \textcolor{red}{(+0.024)}  & 0.551 \textcolor{red}{(+0.510)} \\
\rowcolor{GroupLeft}
+ BA w/ their mask & 0.122 (\textcolor{red}{+0.026}) & 0.032 (\textcolor{red}{+0.018}) & 0.239 (\textcolor{red}{+0.198}) \\
\rowcolor{GroupLeft}
+ BA w/ our mask & 0.096 \textcolor{PineGreen}{(-)} &  0.014 \textcolor{PineGreen}{(-)} & 0.059 \textcolor{red}{(+0.018)} \\

\rowcolor{GroupRight}
AnyCam~\cite{wimbauer2025anycam} & 0.095 & 0.035  & 0.507 \\
\rowcolor{GroupRight}
+ BA w/o mask & 0.123 \textcolor{red}{(+0.028)} & 0.035 \textcolor{PineGreen}{(-)}  & 0.413 \textcolor{PineGreen}{(-0.094)} \\
\rowcolor{GroupRight}
+ BA w/ their mask & 0.098 \textcolor{red}{(+0.003)} & 0.026 \textcolor{PineGreen}{(-0.009)} & 0.365 \textcolor{PineGreen}{(-0.142)} \\
\rowcolor{GroupRight}
+ BA w/ our mask & 0.100 \textcolor{red}{(+0.005)}  & 0.026 \textcolor{PineGreen}{(-0.009)}  & 0.282 \textcolor{PineGreen}{(-0.225)} \\

\rowcolor{GroupLeft}
Monst3r~\cite{zhang2024monst3r} & 0.157 & 0.025 & 0.402 \\
\rowcolor{GroupLeft}
+ BA w/o mask & 0.153 \textcolor{PineGreen}{(-0.004)} & 0.029 \textcolor{red}{(+0.004)} & 0.302 \textcolor{PineGreen}{(-0.100)}  \\
\rowcolor{GroupLeft}
+ BA w/ their mask & 0.140 \textcolor{PineGreen}{(-0.017)} & 0.022 \textcolor{PineGreen}{(-0.003)} & 0.256 \textcolor{PineGreen}{(-0.146)}\\

\rowcolor{GroupLeft}
+ BA w/ our mask & 0.137 \textcolor{PineGreen}{(-0.020)} & 0.021 \textcolor{PineGreen}{(-0.004)} & 0.243 \textcolor{PineGreen}{(-0.159)} \\

\rowcolor{GroupRight}
Easi3r~\cite{chen2025easi3r} & 0.141 & 0.035 & 0.408  \\
\rowcolor{GroupRight}
+ BA w/o mask & 0.148 \textcolor{red}{(+0.007)} &  0.028 \textcolor{PineGreen}{(-0.007)} & 0.293 \textcolor{PineGreen}{(-0.115)} \\
\rowcolor{GroupRight}
+ BA w/ their mask & 0.107 \textcolor{PineGreen}{(-0.034)}  & 0.019\textcolor{PineGreen}{(-0.016)}  & 0.190 \textcolor{PineGreen}{(-0.218)} \\

\rowcolor{GroupRight}
+ BA w/ our mask & 0.092 \textcolor{PineGreen}{(-0.049)} & 0.017 \textcolor{PineGreen}{(-0.018)} & 0.157 \textcolor{PineGreen}{(-0.251)} \\

\bottomrule
\end{tabular}
\label{tab:ba-left_sintel}
\end{minipage}
\hspace{0.13\textwidth} 
\begin{minipage}[t]{0.4\textwidth} 
\centering
\begin{tabular}{@{}l c c c@{}}
\toprule
\multicolumn{4}{@{}c}{\textbf{Methods that do not predict motion masks}} \\
\cmidrule(lr){1-4}
Method & ATE $\downarrow$ & RTE $\downarrow$ & RRE $\downarrow$ \\
\midrule

\rowcolor{GroupLeft}
SpatialTrackerv2~\cite{xiao2025spatialtracker} & 0.095 & 0.022 & 0.120 \\
\rowcolor{GroupLeft}
+ BA w/o mask & 0.141 \textcolor{red}{(+0.046)} & 0.046 \textcolor{red}{(+0.024)} & 0.660 \textcolor{red}{(+0.540)} \\
\rowcolor{GroupLeft}
+ BA w/ our mask & 0.095 \textcolor{PineGreen}{(-)} & 0.022 \textcolor{PineGreen}{(-)} & 0.130 \textcolor{red}{(+0.010)} \\

\rowcolor{GroupRight}
VGGT~\cite{wang2025vggt} & 0.137 & 0.092 & 1.481 \\
\rowcolor{GroupRight}
+ BA w/o mask & 0.130 \textcolor{PineGreen}{(-0.007)} & 0.083 \textcolor{PineGreen}{(-0.009)} & 1.325 \textcolor{PineGreen}{(-0.156)} \\
\rowcolor{GroupRight}
+ BA w/ our mask & 0.130 \textcolor{PineGreen}{(-0.007)} & 0.083 \textcolor{PineGreen}{(-0.009)} & 1.304 \textcolor{PineGreen}{(-0.177)} \\

\rowcolor{GroupLeft}
Cut3r~\cite{Wang2024videocutler} & 0.134 & 0.041 & 0.421 \\
\rowcolor{GroupLeft}
+ BA w/o mask & 0.127 \textcolor{PineGreen}{(-0.007)} & 0.029 \textcolor{PineGreen}{(-0.012)}  & 0.362 \textcolor{PineGreen}{(-0.059)} \\
\rowcolor{GroupLeft}
+ BA w/ our mask & 0.127 \textcolor{PineGreen}{(-0.007)} & 0.029 \textcolor{PineGreen}{(-0.012)} & 0.283 \textcolor{PineGreen}{(-0.138)} \\

\rowcolor{GroupRight}
TTT3r~\cite{chen2025ttt3r} & 0.125 & 0.056 & 0.453 \\
\rowcolor{GroupRight}
+ BA w/o mask & 0.106 \textcolor{PineGreen}{(-0.019)} & 0.037 \textcolor{PineGreen}{(-0.026)}  & 1.036 \textcolor{red}{(+0.583)} \\
\rowcolor{GroupRight}
+ BA w/ our mask & 0.116 \textcolor{PineGreen}{(-0.009)} & 0.037 \textcolor{PineGreen}{(-0.019)} & 0.299 \textcolor{PineGreen}{(-0.154)} \\

\bottomrule
\end{tabular}
\label{tab:ba-right}
\end{minipage}
\vspace{-4mm}
\end{table*}

We report more results of pose optimization in this part, including metrics evaluation on the synthetic dataset Sintel~\cite{Butler:ECCV:2012} and visual comparison on the real-world dataset DAVIS~\cite{Huang_2016}. As shown in Table~\ref{tab:ba-left_sintel}, our BA brings better pose estimation for baselines except MegaSam~\cite{zhang2024megasam} and SpatialTrackerv2~\cite{xiao2025spatialtracker}. Sintel~\cite{Butler:ECCV:2012} is a less challenging dataset compared to LightSpeed~\cite{rockwell2025dynamic}, and these methods are less affected by the dynamic objects. For other baselines that are hindered by dynamic objects, pose optimization with \ourmodel relieves the side effect.   
\section{More Depth Optimization Results}
\begin{figure*}
    \centering
    \includegraphics[width=1\linewidth]{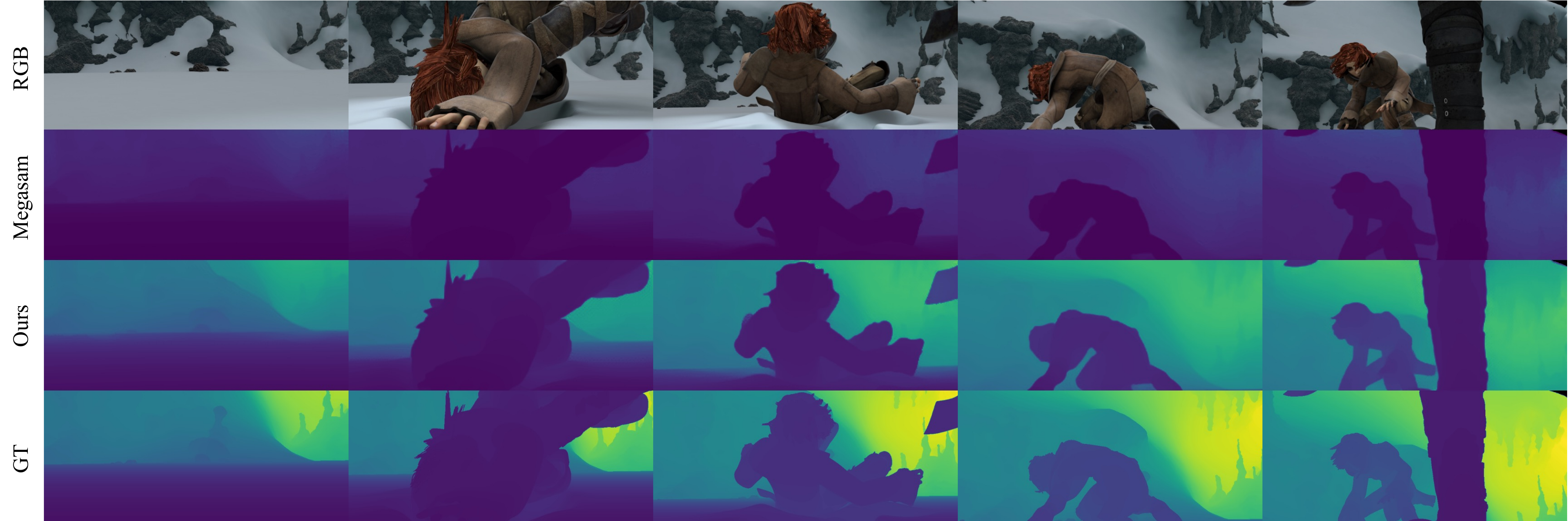}
    \caption{Depth optimization on Sintel~\cite{Huang_2016} ambush\_4.}
    \label{fig:depth_opt_ambush_4}
\end{figure*}

\begin{figure*}
    \centering
    \includegraphics[width=1\linewidth]{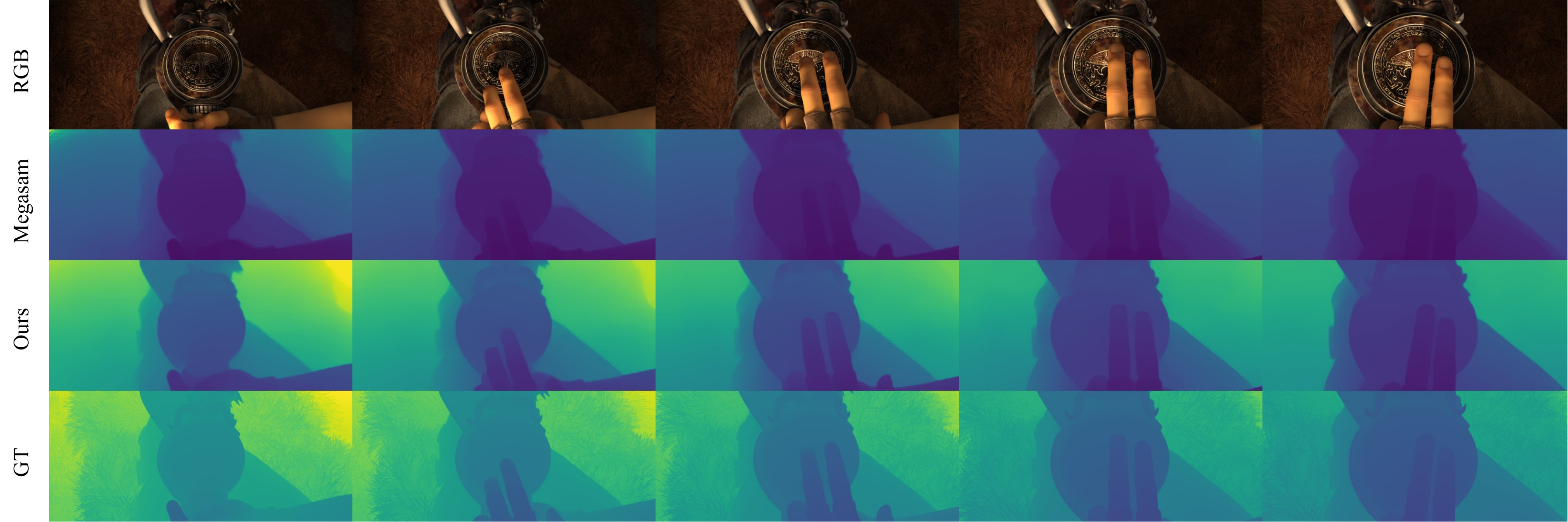}
    \caption{Depth optimization on Sintel~\cite{Huang_2016} shaman\_3.}
    \label{fig:depth_opt_shaman_3}
\end{figure*}

More visual comparison of depth optimization can be found in Fig.~\ref{fig:depth_opt_ambush_4} and Fig.~\ref{fig:depth_opt_shaman_3}. \ \ourmodel generates a better depth of field.

\begin{figure}[!t]
    \vspace{-6mm}
    \centering
    \includegraphics[width=1\linewidth]{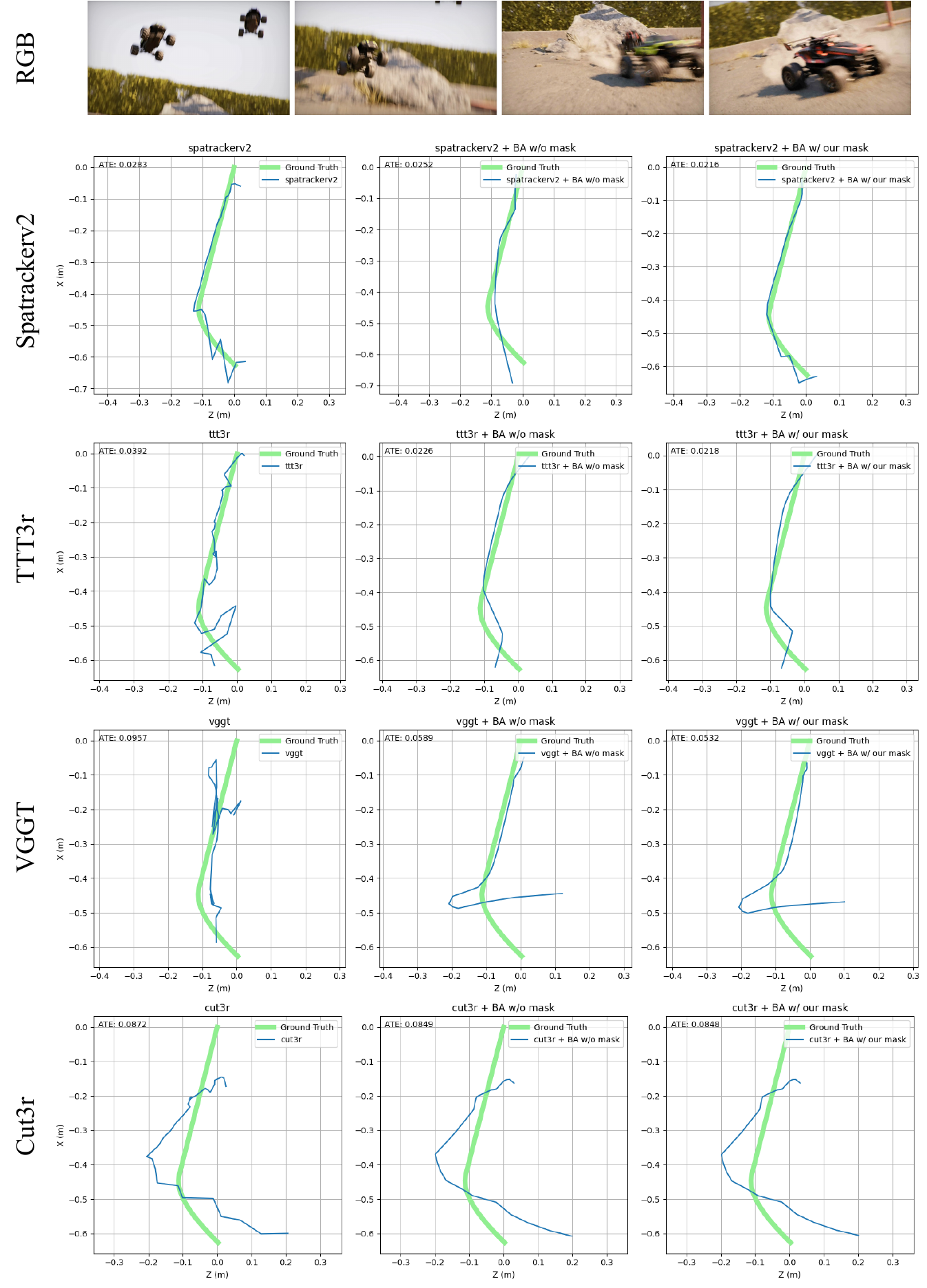}
    \caption{Pose optimization for feed-forward methods on LightSpeed~\cite{rockwell2025dynamic} 0180\_DUST.}
    \label{fig:pose_opt_0180}
    \vspace{-2mm}
\end{figure}

\begin{figure}[!t]
    \centering
    \includegraphics[width=1\linewidth]{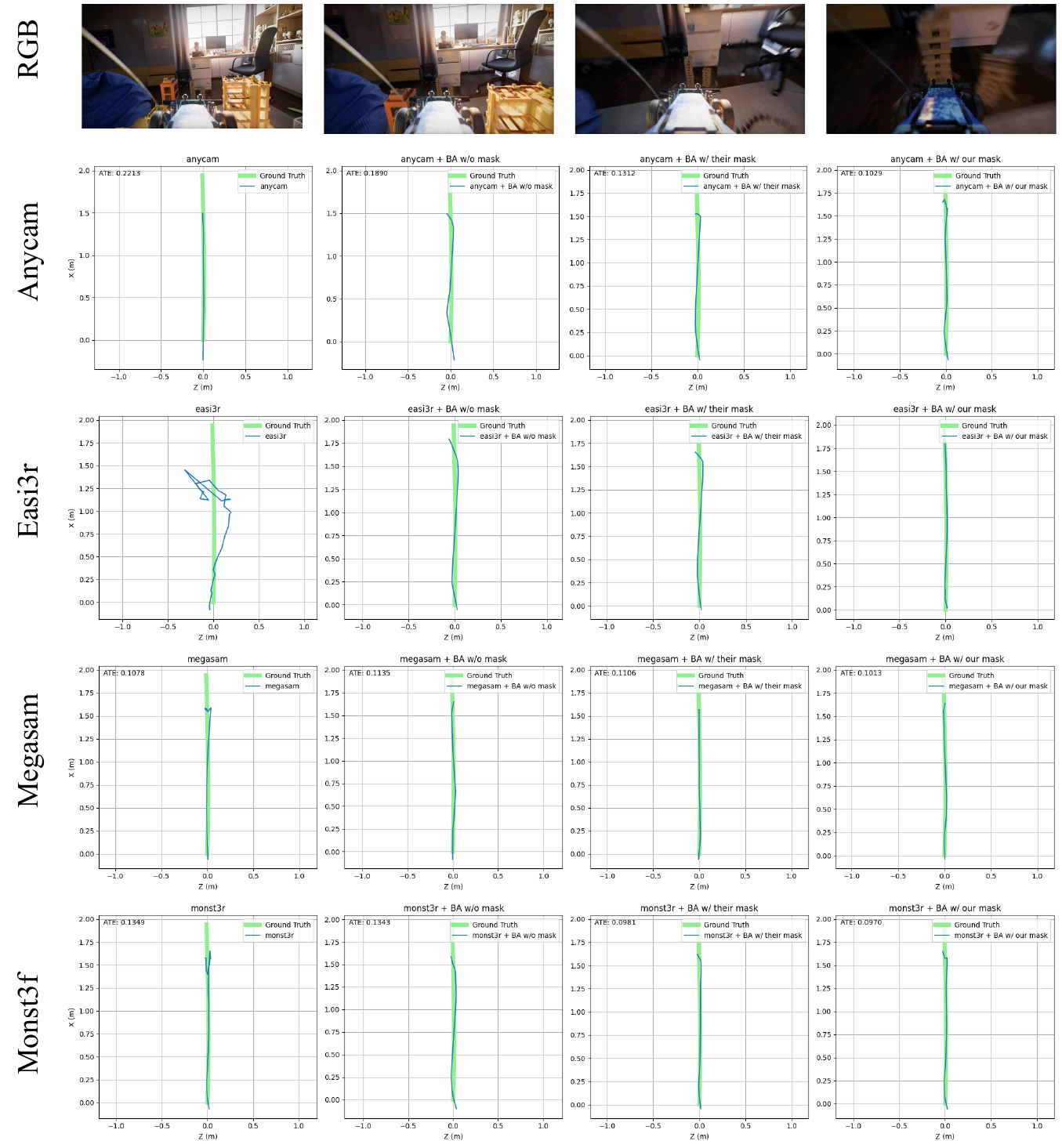}
    \caption{Pose optimization for optimization-based methods on LightSpeed~\cite{rockwell2025dynamic} 0530\_LOFTRAMP.}
    \label{fig:pose_opt_0530}
\end{figure}

\begin{figure}
    \centering
    \includegraphics[width=1\linewidth]{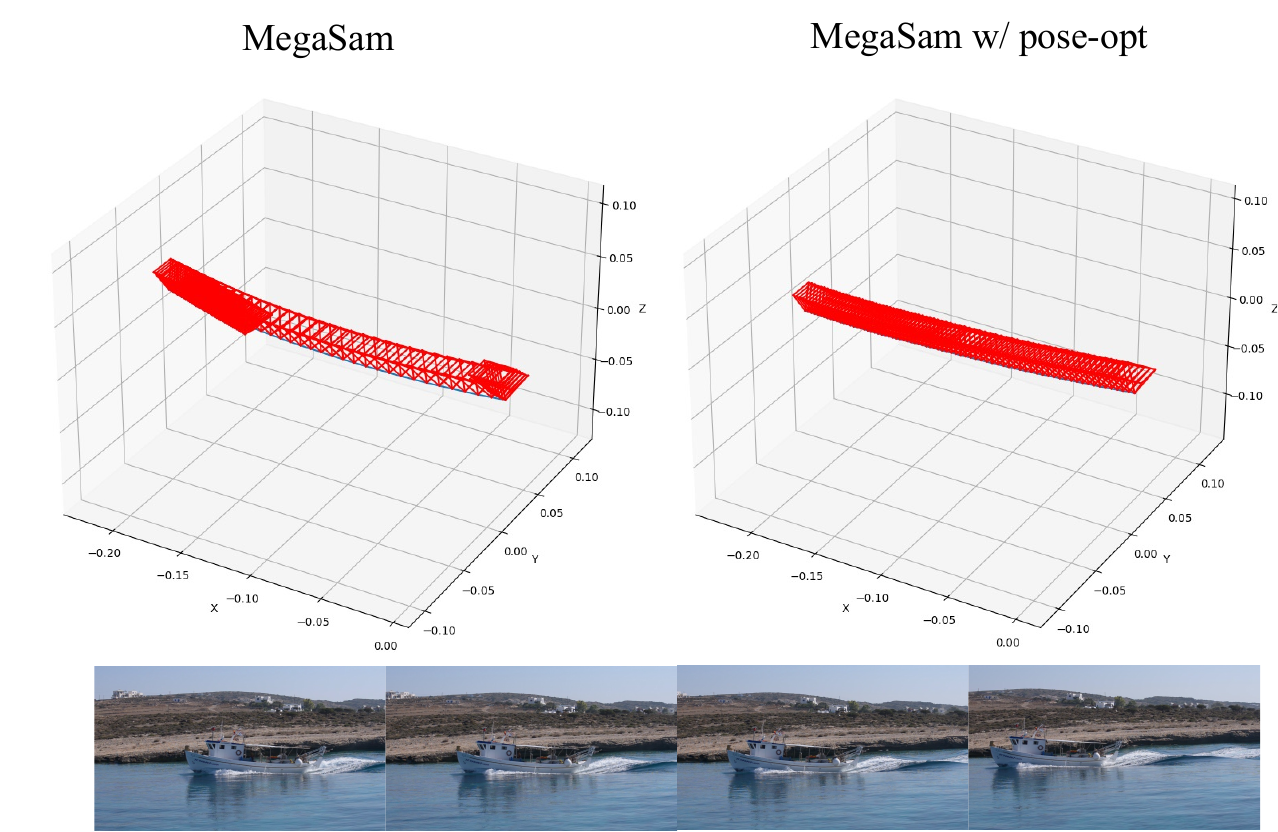}
    \caption{Pose optimization on DAVIS~\cite{Huang_2016} boat.}
    \label{fig:boat}
\end{figure}

\begin{figure}
    \centering
    \includegraphics[width=1\linewidth]{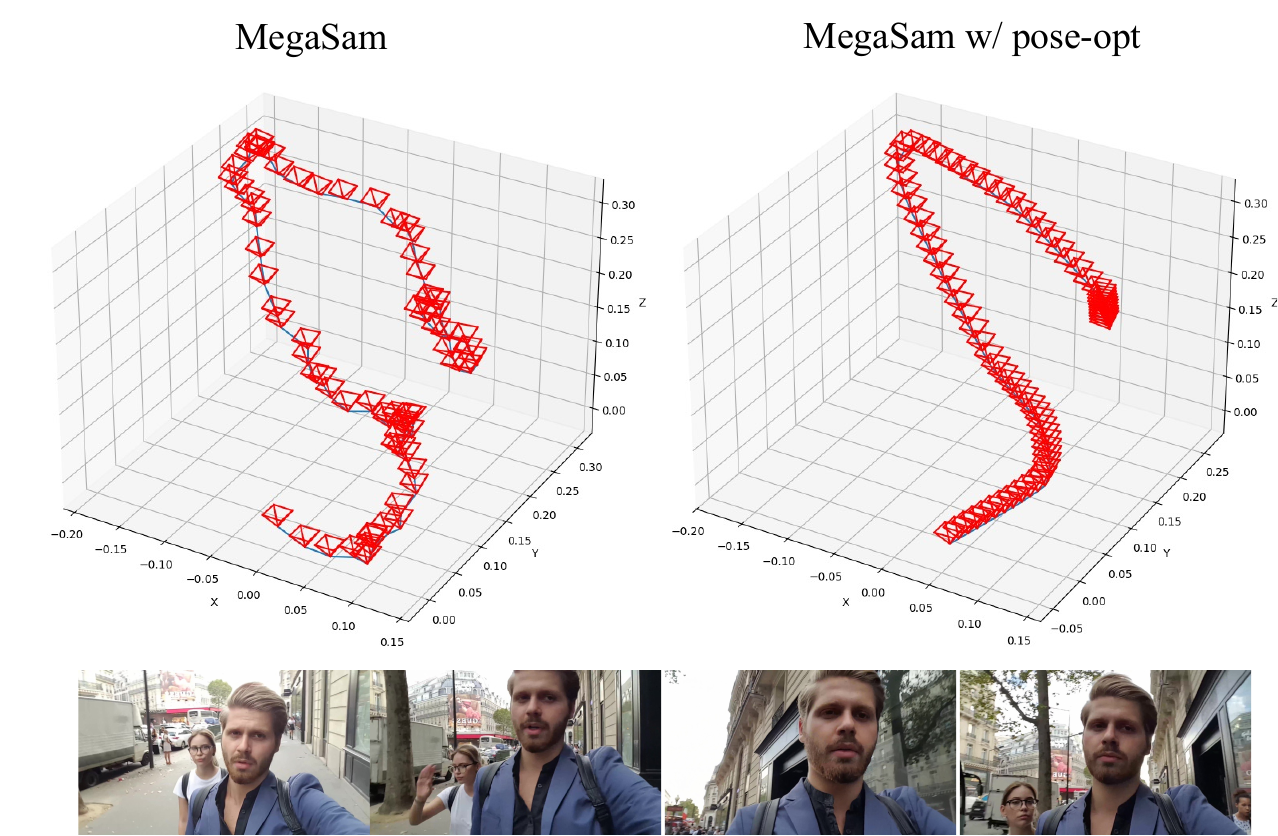}
    \caption{Pose optimization on DAVIS~\cite{Huang_2016} walking.}
    \label{fig:walking}
\end{figure}

\section{More 4D Track Optimization Results}
\begin{figure}[!t]
    \centering
    \includegraphics[width=1\linewidth]{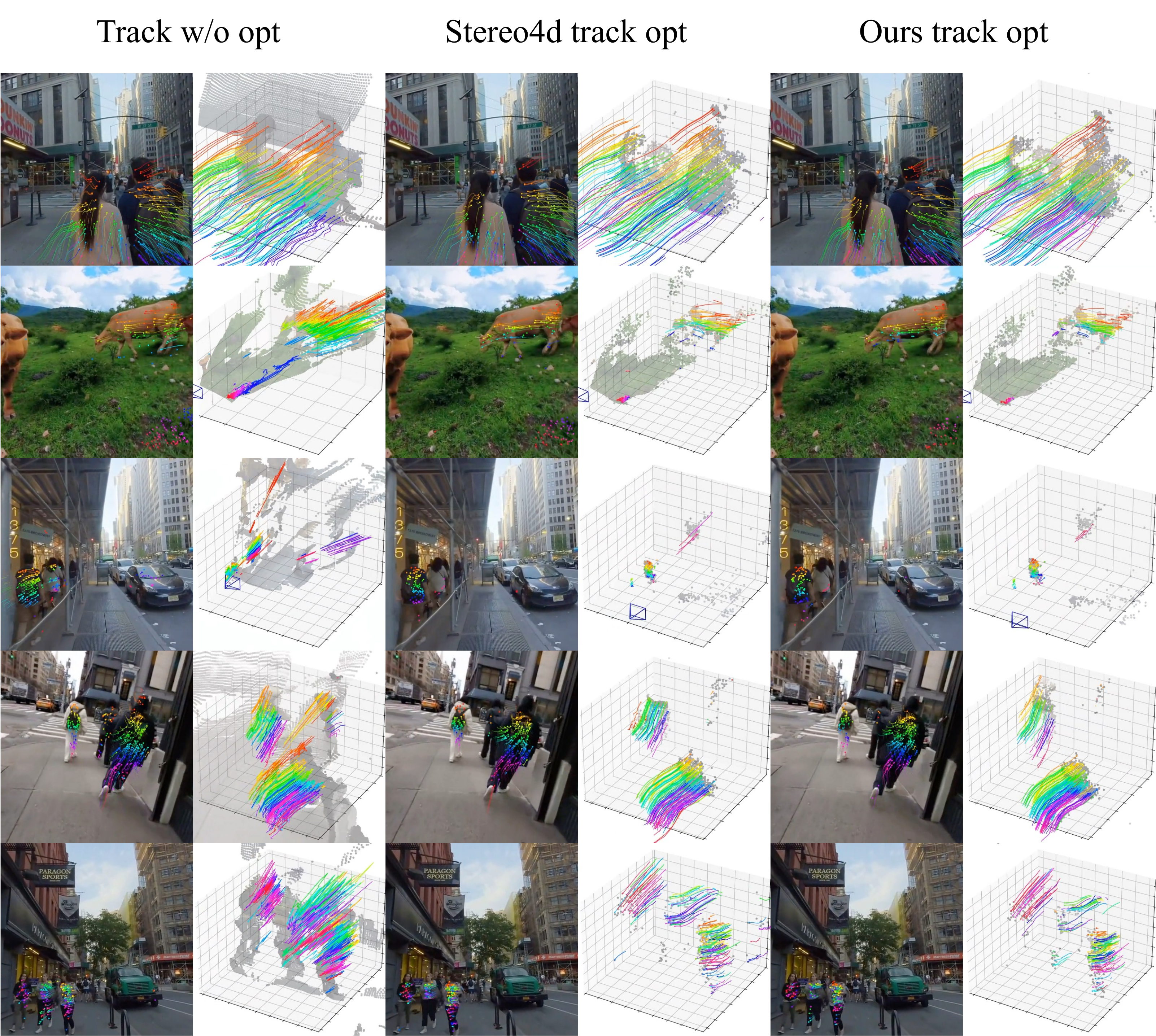}
    \caption{Visual comparison of 4D track optimization of Stereo4D~\cite{jin2025stereo4d} and ours.}
    \label{fig:4d_track_opt}
\end{figure}

We provide more 4D track optimization results in Fig.~\ref{fig:4d_track_opt}. Compared to Stereo4d~\cite{jin2025stereo4d}, there are fewer drifted tracks with our optimization.

\begin{figure}
    \centering
    \includegraphics[width=1\linewidth]{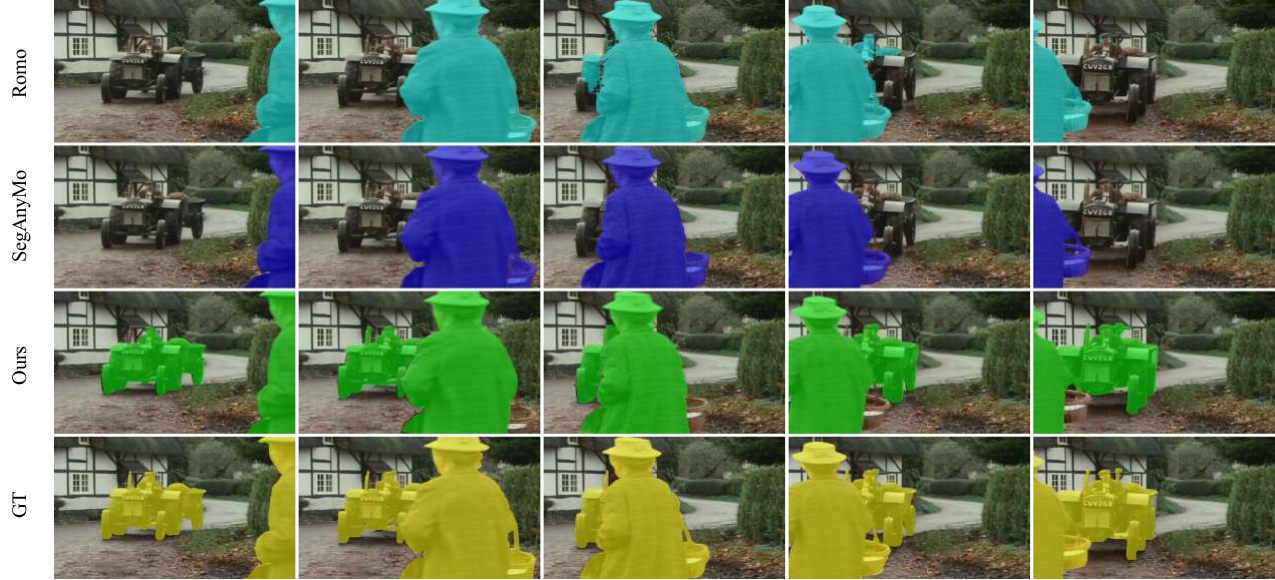}
    \caption{Visual comparison of moving object segmentation on FBMS-59~\cite{ochs2014fbms59} marple8.}
    \label{fig:placeholder}
\end{figure}

\begin{figure}
    \centering
    \includegraphics[width=1\linewidth]{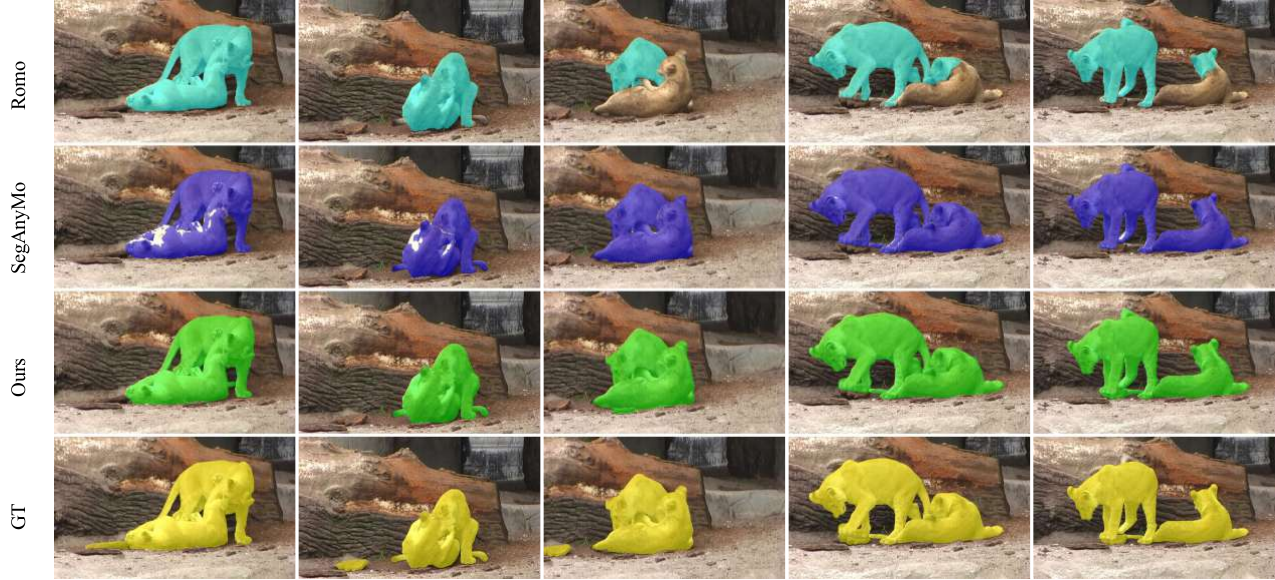}
    \caption{Visual comparison of moving object segmentation on FBMS-59~\cite{ochs2014fbms59} .}
    \label{fig:placeholder}
\end{figure}

\end{document}